\documentclass[letterpaper, 10 pt, journal, twoside]{IEEEtran}
\usepackage[hyphens]{url}
\usepackage{graphicx}
\usepackage{bibentry}
\usepackage{multirow}
\usepackage{array}
\usepackage{float}                                                  
\usepackage{url}            
\usepackage{booktabs}       
\usepackage{tabularx}
\usepackage{amsfonts}       
\usepackage{nicefrac}       
\usepackage{microtype}      
\usepackage{algpseudocode}
\usepackage{algorithm}
\usepackage{algorithmicx}
\usepackage[export]{adjustbox}
\usepackage{makecell}
\usepackage{cite}
\usepackage{arydshln}
\usepackage{caption}
\usepackage[table,dvipsnames]{xcolor}
\usepackage{marvosym}

\definecolor{mediumaquamarine}{rgb}{0.4, 0.8, 0.67}
\newcommand{\rankcell}[2]{{\setlength{\fboxsep}{0.8pt}\colorbox{#1}{\makebox[3.8em][c]{\textcolor{black}{#2}}}}}
\newcommand{\rankone}[1]{\rankcell{red!35}{#1}}
\newcommand{\ranktwo}[1]{\rankcell{orange!35}{#1}}
\newcommand{\rankthree}[1]{\rankcell{yellow!55}{#1}}
\newcommand{\ranklegend}{{\setlength{\fboxsep}{0.3pt}\colorbox{red!25}{\textcolor{black}{1\textsuperscript{st}}} / \colorbox{orange!35}{\textcolor{black}{2\textsuperscript{nd}}} / \colorbox{yellow!55}{\textcolor{black}{3\textsuperscript{rd}}}}}

\newcommand{\eqnref}[1]{Eq.~(\ref{#1})}

\usepackage{newfloat}
\usepackage{listings}
\usepackage{svg}
\usepackage{amsmath}
\usepackage{bm}
\usepackage{capt-of}

\newcolumntype{M}[1]{>{\centering\arraybackslash}m{#1}}

\usepackage{hyperref}
\hypersetup{
  colorlinks=true,
  urlcolor=RoyalBlue,
  linkcolor=black,
  citecolor=black,
  breaklinks=true
}

\IEEEoverridecommandlockouts
\title{\textbf{Swimm3R}: \textbf{S}platting \textbf{wi}th \textbf{M}edium-aware Sf\textbf{M} for Underwater \textbf{3}D \textbf{R}econstruction}
\author{Minseong Kweon$^{1}$ and Junaed Sattar$^{1}$
\\[8pt]
{\normalfont\normalsize\Mundus\ \textbf{Project Page:} \href{https://mnseong.github.io/swimm3r.github.io/}{\textbf{Link}}}%
\vspace{-10pt}
\thanks{$^{1}$The authors are with the Department of Computer Science \& Engineering
and the Minnesota Robotics Institute (MnRI), University of
Minnesota--Twin Cities, Minneapolis, MN 55455 USA
{\tt\footnotesize \{kweon021, junaed\}@umn.edu}
}
}

\makeatletter
\g@addto@macro\normalsize{%
  \setlength\abovedisplayskip{5pt plus 1pt minus 1pt}%
  \setlength\belowdisplayskip{5pt plus 1pt minus 1pt}%
  \setlength\abovedisplayshortskip{3pt plus 1pt}%
  \setlength\belowdisplayshortskip{3pt plus 1pt}%
}
\makeatother

\begin{document}
\sloppy

\maketitle

\begin{abstract}
We propose Swimm3R, a unified framework that combines medium-aware structure-from-motion (SfM) with Underwater Beta Splatting to address scattering- and attenuation-induced failures in underwater 3D reconstruction.
Swimm3R distills in-air geometric priors into a feed-forward backbone and uses a physics head to regress underwater image-formation parameters, camera poses, and restored point clouds.
Additionally, we introduce Underwater Beta Splatting, which extends Gaussian splatting with Beta primitives and scattering-aware geometric gradients for stable underwater geometry representation.
We further establish the Barbados underwater video dataset to demonstrate the effectiveness of our method in challenging underwater environments.
On this dataset, Swimm3R robustly recovers underwater scene structure under challenging scattering conditions, yielding coherent seafloor geometry.
Using these predicted point clouds, the proposed Underwater Beta Splatting improves average PSNR by $1.47$ dB over WaterSplatting while increasing downstream localization performance by $2.0$ and $2.4$ percentage points in RRA@15 and RTA@15, respectively.
\end{abstract}

\begin{IEEEkeywords}
Deep Learning for Visual Perception, Marine Robotics, Seafloor Mapping.
\end{IEEEkeywords}

\section{Introduction}

\IEEEPARstart{U}{nderwater} 3D reconstruction underpins marine robotic localization, navigation, and human–robot interaction~\cite{wang2023real}.
For visual localization and SLAM, reconstructed 3D maps must be geometrically consistent and visually faithful, as errors in geometry or rendering degrade feature matching and pose estimation.
However, underwater imaging suffers from wavelength-dependent attenuation and backscatter, which degrade color, contrast, and scene visibility~\cite{kaeli2017illumination,akkaynak2018revised}.

To improve underwater image quality, prior work has extensively studied restoration and enhancement.
Early methods studied underwater image formation models of attenuation and backscatter~\cite{mcglamery1980computer,jaffe1990computer} with hand-crafted priors, such as dark-channel, red-channel, and transmission estimation, to recover color and contrast~\cite{he2010single,galdran2015automatic}.
These models were later refined with range- and wavelength-dependent optical properties~\cite{akkaynak2018revised,akkaynak2019sea}.
More recently, learning-based methods formulate restoration as a data-driven mapping from degraded observations to enhanced targets~\cite{islam2020fast}.
Beyond image-level restoration, neural 3D representations such as Neural Radiance Fields (NeRF)~\cite{mildenhall2020nerf} and 3D Gaussian Splatting (3DGS)~\cite{kerbl20233d} jointly optimize scene geometry and appearance with physics-grounded underwater image formation models~\cite{akkaynak2019sea}. Accordingly, recent underwater frameworks~\cite{tang2024neural,levy2023seathru,li2024watersplatting,yang2024seasplat,wu2025plenodium,qiao2025restorgs,kweon2026oceansplat} represent underwater scene structure together with scattering-corrected appearance.
These methods, however, rely on SfM-derived poses and sparse points, which become unreliable in highly turbid or dynamic scenes where scattering and attenuation degrade sparse correspondences.

\begin{figure}[t]
\centering
\includegraphics[width=0.93\columnwidth]{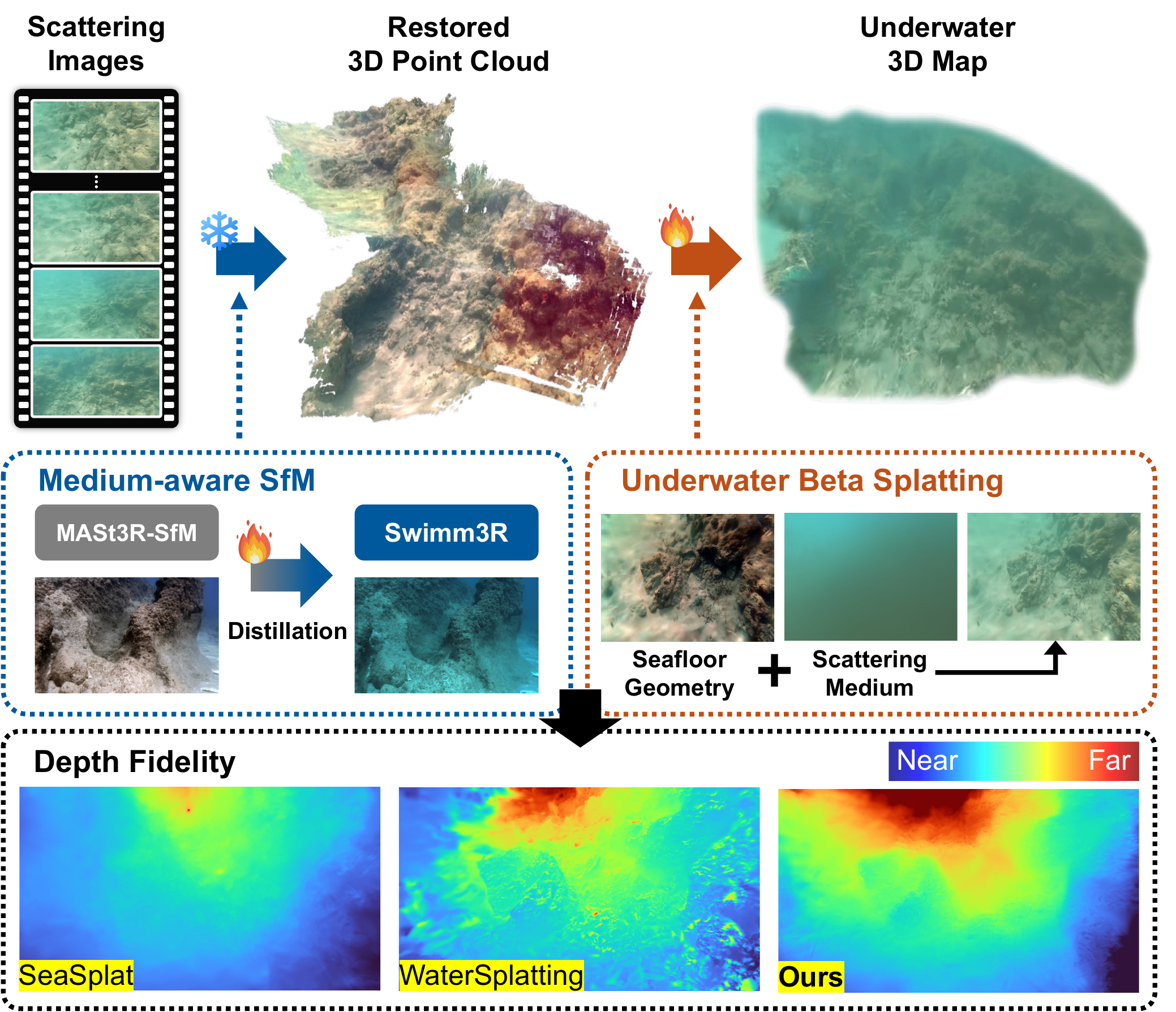}
\caption{Swimm3R achieves medium-aware SfM through in-air to underwater geometry distillation, and UWBS uses this prior for realistic underwater scene reconstruction.}
\label{fig:teaser}
\end{figure}

To overcome these limitations, we propose \textbf{Swimm3R}, a unified framework that couples Underwater Beta \textbf{S}platting \textbf{wi}th a \textbf{m}edium-aware structure-from-\textbf{m}otion for underwater \textbf{3}D \textbf{R}econstruction.
The medium-aware SfM pipeline fine-tunes MASt3R-SfM~\cite{duisterhof2025mast3r} with LoRA~\cite{hu2022lora} to transfer its robust in-air matching and geometric priors to underwater imagery.
In addition, we introduce an underwater physics head that estimates optical parameters in a feed-forward manner, enabling the recovery of restored dense 3D point clouds at inference time.
We further introduce Underwater Beta Splatting (UWBS), which models backscatter and wavelength-dependent attenuation with an explicit underwater medium field and uses generalized Beta kernels~\cite{liu2025universal} to jointly represent geometry, view-dependent appearance, and static and dynamic underwater scenes.
To optimize Beta primitives under scattering, we design Scattering-aware Geometric Gradients, which suppress medium-dominated projected-center updates and recover geometry-supported primitive motion.
Fig.~\ref{fig:teaser} qualitatively shows more coherent geometry and restored appearance than prior methods.
Moreover, to evaluate Swimm3R in challenging underwater environments, we develop a dataset comprising four GoPro videos captured in Barbados, covering diverse and challenging underwater conditions, as shown in Tab.~\ref{tab:barbados_dataset}.
Our main contributions are summarized as follows:
\begin{itemize}
    \item We propose \textbf{Swimm3R}, a medium-aware SfM framework that adapts in-air priors to underwater imagery and predicts poses, optical parameters, and restored point clouds.
    \item We introduce Underwater Beta Splatting with Scattering-aware Geometric Gradients for joint Beta-primitive and medium-field optimization in underwater scenes.
    \item We establish the Barbados underwater video dataset and demonstrate improvements in geometry estimation, test-view rendering, and downstream visual localization.
\end{itemize}
\section{Related Work}

\subsection{Underwater 3D Computer Vision}

Early systems extended classical SfM~\cite{schonberger2016structure} and MVS~\cite{schoenberger2016mvs} to underwater mapping~\cite{beall20103d}. Prior work further improved underwater geometry through refractive camera models~\cite{jordt2012refractive,she2024refractive} and color correction for feature matching~\cite{skinner2017automatic}.

Recent underwater 3D reconstruction methods integrate underwater image
formation~\cite{akkaynak2018revised,akkaynak2019sea} into Neural Radiance Fields (NeRF)~\cite{mildenhall2020nerf}
and 3D Gaussian Splatting (3DGS)~\cite{kerbl20233d}.
NeRF-based approaches~\cite{sethuraman2023waternerf,levy2023seathru,tang2024neural} jointly model scene geometry, radiance, and medium effects, while 3DGS-based methods~\cite{yang2024seasplat,li2024watersplatting,zhang2024recgs,kweon2026oceansplat,qiao2025restorgs,wu2025plenodium} leverage explicit 3D Gaussian representations for faster optimization and more efficient rendering, making them particularly well suited for geometry-aware underwater reconstruction.
These methods jointly model restoration, medium-aware rendering, geometry, or uncertainty, but many still rely on SfM-derived camera poses and sparse geometric initialization, which remain fragile under severe scattering and low-texture.

In contrast, our proposed method distills a feed-forward model pretrained on in-air imagery to the underwater domain, directly regressing 3D point clouds, camera poses, and underwater optical parameters from uncalibrated underwater videos.

\subsection{Feed-forward 3D Reconstruction}
Feed-forward 3D reconstruction replaces the sequential matching, triangulation, and optimization stages of classical SfM~\cite{schonberger2016structure} and MVS~\cite{schoenberger2016mvs} with learned networks that directly predict scene geometry and camera parameters from images.
DUSt3R~\cite{wang2024dust3r} directly regresses dense pointmaps from uncalibrated image pairs, providing a unified feed-forward formulation for geometry, correspondence, and camera estimation.
Its strong geometric priors have subsequently inspired a broad range of feed-forward reconstruction methods.
MASt3R~\cite{leroy2024grounding} expands DUSt3R with 3D-grounded dense matching, which MASt3R-SfM~\cite{duisterhof2025mast3r} further leverages for robust SfM over unconstrained image collections. Dark3R~\cite{guo2026dark3r} extends this paradigm to extreme low-light imagery.
To avoid costly global alignment, Spann3R~\cite{wang20253d} and VGGT~\cite{wang2025vggt} jointly exploit multiple views in a common scene representation.
Complementary approaches either couple feed-forward geometric priors with subsequent 3DGS~\cite{kerbl20233d} to refine scene representations and camera poses~\cite{fan2024instantsplat,huang20253r}, or integrate them into  global SfM to improve reconstruction robustness and scalability~\cite{pan2026global}.

Motivated by Dark3R~\cite{guo2026dark3r}, our approach adapts feed-forward geometric priors to scattering media through paired underwater–clean supervision, followed by joint refinement of geometry and camera poses with 3DGS~\cite{kerbl20233d}.

\begin{table}[t]
\centering
\caption{Overview of the Barbados underwater video dataset.}
\label{tab:barbados_dataset}
\setlength{\tabcolsep}{1pt}
\renewcommand{\arraystretch}{1.2}
\newlength{\thumbheight}
\setlength{\thumbheight}{0.105\textwidth}
\resizebox{\linewidth}{!}{%
{\Large
\begin{tabular}{l|@{\hspace{2pt}}c@{\hspace{2pt}}c@{\hspace{2pt}}c@{\hspace{2pt}}c@{\hspace{2pt}}}
Scenes
& Cyan & Murky & Outcrop & Caustics \\
\hline
\multirow{3}{*}{Thumbnail}
& \multirow{3}{*}{\includegraphics[height=\thumbheight,keepaspectratio]{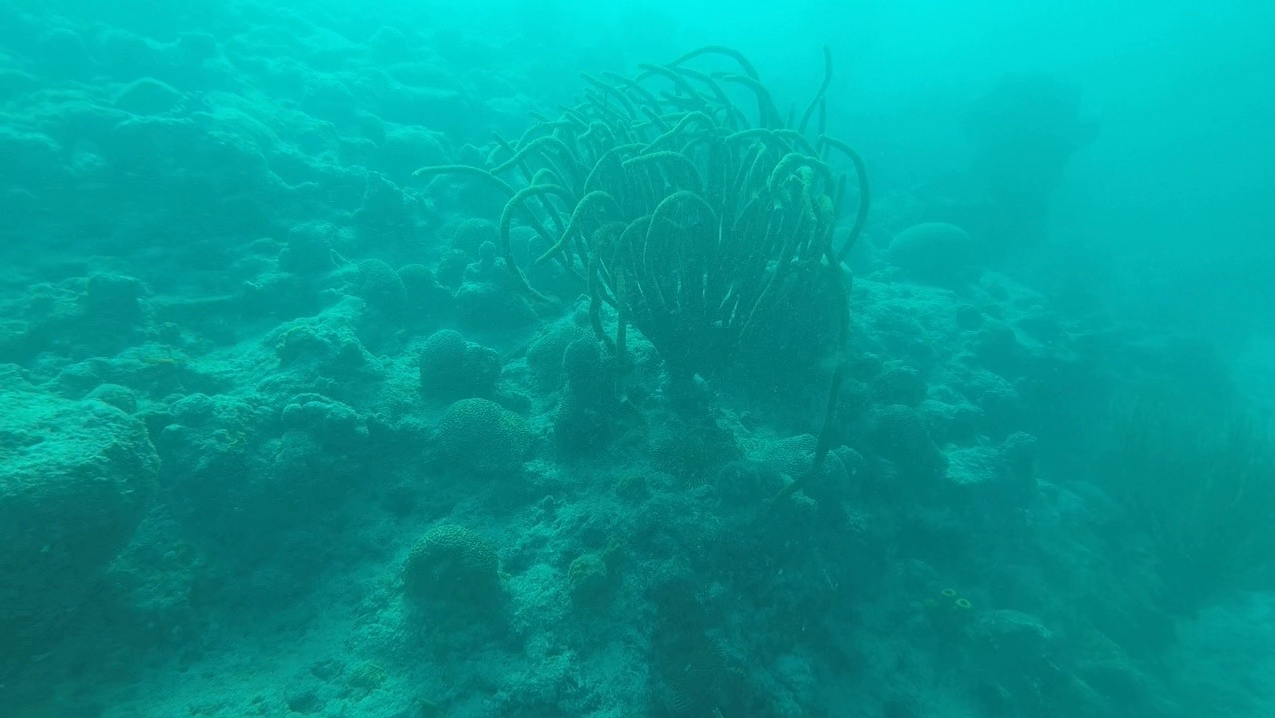}}
& \multirow{3}{*}{\includegraphics[height=\thumbheight,keepaspectratio]{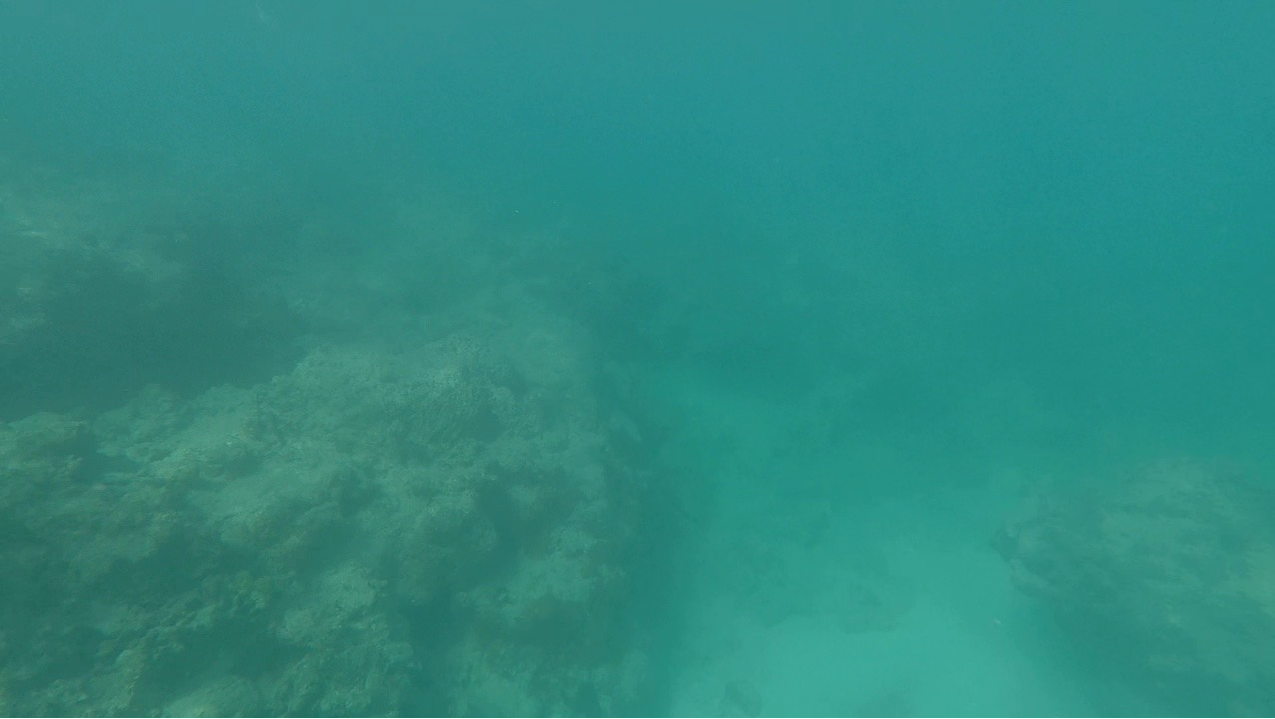}}
& \multirow{3}{*}{\includegraphics[height=\thumbheight,keepaspectratio]{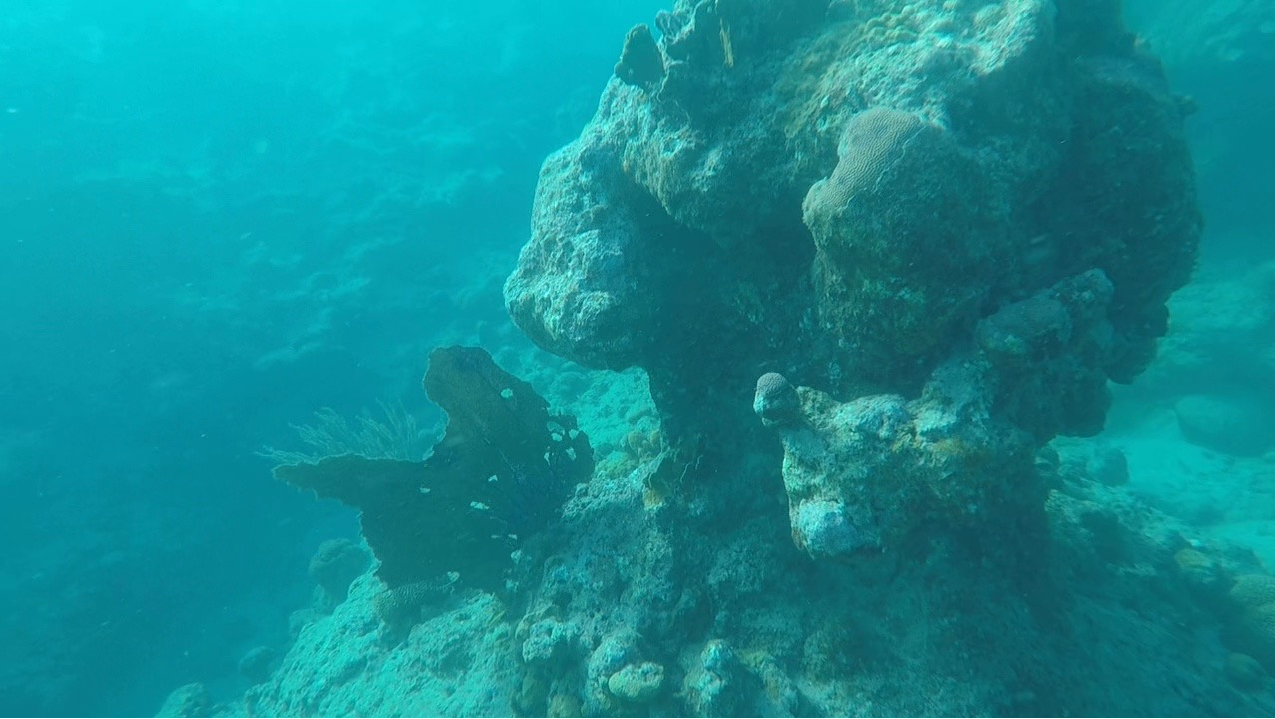}}
& \multirow{3}{*}{\includegraphics[height=\thumbheight,keepaspectratio]{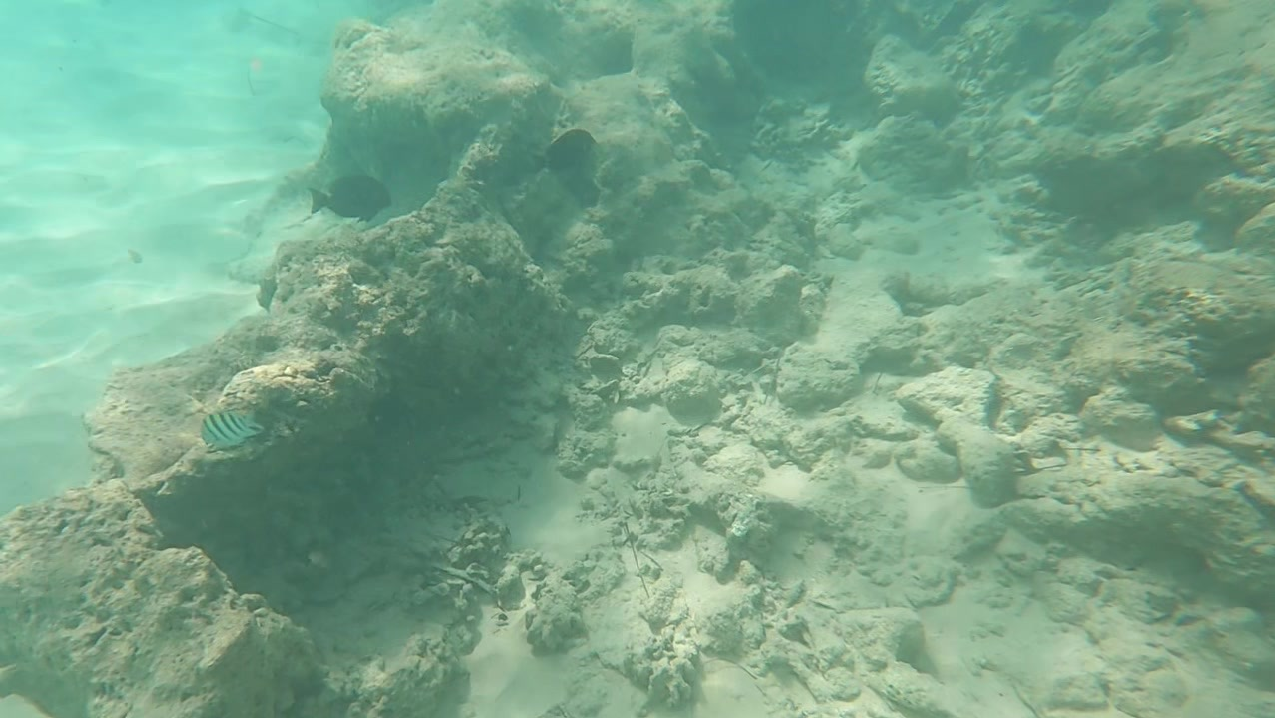}} \\
& & & & \\
& & & & \\
\hdashline
Condition
& Cyan veiling & Turbid water & Seafloor outcrop & Caustic flicker \\
\hdashline
Length (sec.)
& 23 & 30 & 16 & 25 \\
\hdashline
Resolution
& \multicolumn{4}{c}{$1280\times720$} \\
\hdashline
FPS (raw/sampled)
& \multicolumn{4}{c}{$30 \,/\, 5$} \\
\hdashline
Frames (train/test)
& $102 \,/\, 15$ & $131 \,/\, 19$ & $71 \,/\, 11$ & $109 \,/\, 16$ \\
\hline
\end{tabular}
}
}%
\end{table}

\begin{figure*}[t]
    \centering
    \includegraphics[width=0.85\linewidth]{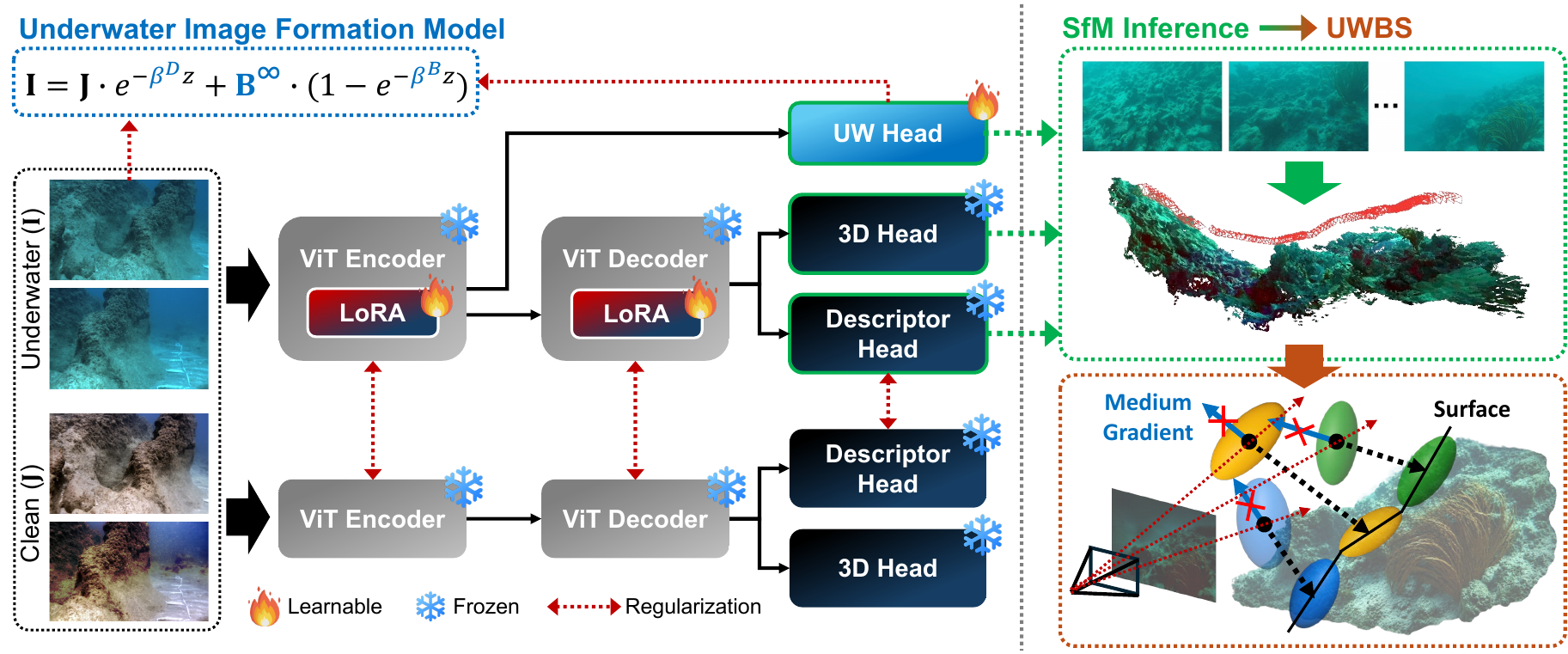}
    \caption{\textbf{Overview of Swimm3R.} Our model learns medium-aware SfM from raw underwater images via LoRA fine-tuning, whose estimated camera poses and point clouds provide rich geometric priors for underwater Beta splatting.}
    \label{fig:overview}
\end{figure*}

\section{Swimm3R}
\subsection{Overview}
\label{sec:swimm3r_overview}

Fig.~\ref{fig:overview} illustrates the overall pipeline of Swimm3R.
We propose medium-aware SfM, which distills the strong in-air geometric prior of MASt3R-SfM~\cite{duisterhof2025mast3r} into the underwater domain, enabling robust feed-forward feature matching under scattering conditions. A lightweight physics head predicts direct attenuation $\boldsymbol{\beta}^{D}$,
backscatter $\boldsymbol{\beta}^{B}$, and veiling light
$\mathbf{B}^{\infty}$ from raw underwater images, from which dense geometry and
restored point colors are inferred for initializing Beta primitives for underwater 3D reconstruction.

\subsection{Medium-aware Structure-from-Motion}
\label{sec:medium_aware_sfm}

\noindent \textbf{Underwater SfM and medium regression.}
Following the LoRA~\cite{hu2022lora} fine-tuning strategy of Dark3R~\cite{guo2026dark3r}, we distill clean-image geometric priors into raw underwater imagery.
We distill the teacher's encoder, decoder, and descriptor feature maps, denoted by $\mathbf{F}_{\mathrm{enc}}, \mathbf{F}_{\mathrm{dec}}, \mathbf{F}_{\mathrm{desc}}$, into the corresponding student features $\tilde{\mathbf{F}}_{\mathrm{enc}}, \tilde{\mathbf{F}}_{\mathrm{dec}}, \tilde{\mathbf{F}}_{\mathrm{desc}}$ using
\begin{equation}
\mathcal{L}_{\mathrm{geom}} =
\sum_{k\in\{\mathrm{enc},\mathrm{dec},\mathrm{desc}\}}
\lambda_k
\left\lVert
\mathbf{F}_k - \tilde{\mathbf{F}}_k
\right\rVert_2^2,
\label{eq:geom}
\end{equation}
where $\lambda_{\mathrm{enc}}=\lambda_{\mathrm{dec}}=0.5$ and $\lambda_{\mathrm{desc}}=2.0$.
A lightweight underwater head parameterized by the student encoder features evaluates its per-view coefficients at the metric ray depth $z$.
This head predicts a Sea-Thru~\cite{akkaynak2019sea} depth-dependent direct attenuation map $\boldsymbol{\beta}^{D}(z)$, a depth-independent backscatter coefficient $\boldsymbol{\beta}^{B}$ that is broadcast over pixels, and frame-level veiling light $\mathbf{B}^{\infty}$.

\noindent \textbf{Underwater image formation model in SfM.}
Let $\mathbf{I}_n$ denote a raw underwater pixel and $\mathbf{J}_n$ its paired restored target. For the $n^{th}$ valid pixel at metric ray depth $z_n$, define the direct and backscatter transmittances
\begin{equation}
\mathbf{T}^{D}_{n}=e^{-\boldsymbol{\beta}^{D}_{n}z_n},
\qquad
\mathbf{T}^{B}_{n}=e^{-\boldsymbol{\beta}^{B}z_n}.
\label{eq:stage1_trans}
\end{equation}
The physics head defines a forward underwater rendering path and an inverse
restoration path,
\begin{equation}
\mathbf{I}^{\rightarrow}_{n}
=
\mathbf{J}_{n}\odot\mathbf{T}^{D}_{n}
+
\mathbf{B}^{\infty}\odot(1-\mathbf{T}^{B}_{n}),
\label{eq:swimm3r_render}
\end{equation}
\begin{equation}
\tilde{\mathbf{J}}^{\leftarrow}_{n}
=
\frac{\mathbf{I}_{n}-\mathbf{B}^{\infty}\odot(1-\mathbf{T}^{B}_{n})}
{\max(\mathbf{T}^{D}_{n},\epsilon_{\mathrm{inv}})},
\qquad
\hat{\mathbf{J}}_{n}
=
\left[\tilde{\mathbf{J}}^{\leftarrow}_{n}\right]_{0}^{1}.
\label{eq:swimm3r_render_inverse}
\end{equation}
The physics head is supervised by forward and inverse losses:
\begin{equation}
\mathcal{L}_{\mathrm{for}}
=
\frac{1}{N}\sum_{n=1}^{N}
\left\|\mathbf{I}^{\rightarrow}_{n}-\mathbf{I}_{n}\right\|_{1},
\label{eq:stage1_loss_for}
\end{equation}

\begin{equation}
\mathcal{L}_{\mathrm{inv}}
=
\frac{1}{N}\sum_{n=1}^{N}
\left\|\tilde{\mathbf{J}}^{\leftarrow}_{n}-\mathbf{J}_{n}\right\|_{1}.
\label{eq:stage1_loss_inv}
\end{equation}
To stabilize restoration and avoid red or cyan saturation, we penalize
out-of-range inverse radiance and anchor the veiling light to dim-pixel color
statistics:
\begin{equation}
\begin{aligned}
\mathcal{L}_{\mathrm{sat}}
&=
\frac{1}{N}\sum_{n=1}^{N}
\left(
\left\|[\tilde{\mathbf{J}}^{\leftarrow}_{n}-1]_{+}\right\|_{2}^{2}
+
\left\|[-\tilde{\mathbf{J}}^{\leftarrow}_{n}]_{+}\right\|_{2}^{2}
\right)
\\
&\quad
+
0.5
\left\|
\mathbf{B}^{\infty}
-
Q_{0.05}\!\left(\{\mathbf{I}_{n}\}_{n=1}^{N}\right)
\right\|_{2}^{2}.
\end{aligned}
\label{eq:stage1_loss_sat}
\end{equation}
Here, $N$ is the number of valid pixels, $[\cdot]_{+}$ denotes positive
clamping, and $Q_{0.05}$ is the per-channel bottom-$5\%$ raw-image intensity.
The total SfM training objective is
\begin{equation}
\mathcal{L}_{\mathrm{SfM}}
=
\mathcal{L}_{\mathrm{geom}}
+
\mathcal{L}_{\mathrm{for}}
+
\mathcal{L}_{\mathrm{inv}}
+
\mathcal{L}_{\mathrm{sat}}.
\label{eq:stage1_objective}
\end{equation}
At inference, we apply \eqnref{eq:swimm3r_render_inverse} using the predicted depth and medium parameters, and average the recovered radiance over SfM tracks to obtain a restored point cloud.

\noindent \textbf{Parametric water-type augmentation.}
Estimating $\mathbf{B}^{\infty}$ from a single underwater image is ambiguous due to varying water colors.
We regularize this ambiguity with underwater image augmentation during SfM training.
With probability $p{=}0.5$, we forward-render $\mathbf{J}$ using \eqnref{eq:swimm3r_render} with $\mathbf{B}^{\infty}$ uniformly sampled from four categories:

{\footnotesize
\begin{equation}
\mathbf{B}^{\infty}(R,G,B) \sim
\begin{cases}
([0.02,0.08],[0.30,0.50],[0.30,0.45]) & \text{cyan},\\
([0.05,0.15],[0.45,0.70],[0.30,0.55]) & \text{teal},\\
([0.10,0.25],[0.30,0.50],[0.10,0.30]) & \text{green},\\
([0.20,0.40],[0.15,0.30],[0.10,0.25]) & \text{warm}.
\end{cases}
\end{equation}
}
Here, each interval denotes a uniform sampling range for the corresponding
RGB channel.
We sample $\beta^{D}_{R}\sim\mathcal{U}(0.3,2.0)$, using
$\mathcal{U}(0.7,1.6)$ for the cyan family to target the cyan-dead regime,
and sample $\beta^{D}_{G}$ and $\beta^{D}_{B}$ from $\mathcal{U}(0.04,0.85)$.
Finally, we set $\boldsymbol{\beta}^{B}=r\boldsymbol{\beta}^{D}$, where
$r\sim\mathcal{U}(0,1)$ per channel.

\subsection{Underwater Beta Splatting}
\noindent \textbf{Scene representation.}
\label{sec:scene}
We represent the scene with Beta primitives following Universal Beta Splatting (UBS)~\cite{liu2025universal}. Beta primitives provide adaptive spatial and angular support, allowing them to capture sharp geometry and view-dependent appearance more flexibly than fixed Gaussian kernels. This expressiveness is particularly useful for underwater scenes, where scattering, attenuation, specular highlights, and water-column effects make object boundaries and radiance strongly view-dependent. Let $\mathbf{u}$ denote a pixel location on the image plane and $\mathbf{d}_i$ denote the viewing direction of primitive $i$. Each conditioned primitive contributes opacity $\alpha_i(\mathbf{u})$, color $\mathbf{c}_i$, and depth $z_i$ to the rasterizer. For the water column, we generalize the static medium MLP of WaterSplatting~\cite{li2024watersplatting} with an ambient convolutional field $\mathcal{A}_{\phi}$:
\begin{equation}
\bigl(\mathbf{B}^{\infty},\boldsymbol{\beta}^{B},\boldsymbol{\beta}^{D}\bigr)(\mathbf{u})
=
\sigma_{\mathrm{med}}\!\left[
\mathcal{A}_{\phi}\!\left(
\mathrm{SH}(\mathbf{d}(\mathbf{u}))
\right)
\right],
\label{eq:ambient_medium}
\end{equation}
where $\mathrm{SH}$ is the spherical-harmonic encoding of the pixel ray direction $\mathbf{d}(\mathbf{u})$, and $\mathcal{A}_{\phi}$ is a $1{\times}1$ convolutional network with ReLU hidden activations. The output activation $\sigma_{\mathrm{med}}$ applies sigmoid to $\mathbf{B}^{\infty}$ and softplus with bias to $(\boldsymbol{\beta}^{B},\boldsymbol{\beta}^{D})$.
Given $K$ visible primitives sorted by depth, the fused underwater rasterizer computes
\begin{equation}
\mathbf{V}(\mathbf{u})
=
\sum_{i=1}^{K}
T_i
\left(
e^{-\boldsymbol{\beta}^{B}z_{i-1}}
-
e^{-\boldsymbol{\beta}^{B}z_i}
\right)
+
T_{K+1}e^{-\boldsymbol{\beta}^{B}z_K},
\label{eq:ugs_veil}
\end{equation}
\begin{equation}
\hat{\mathbf{I}}(\mathbf{u})
=
\sum_{i=1}^{K}
T_i\alpha_i(\mathbf{u})\mathbf{c}_i
\odot e^{-\boldsymbol{\beta}^{D}z_i}
+
\mathbf{B}^{\infty}\odot \mathbf{V}(\mathbf{u}).
\label{eq:ugs_render}
\end{equation}
where $T_i=\prod_{j<i}(1-\alpha_j(\mathbf{u}))$, $z_0=0$, $K$ is the last visible primitive, and all medium parameters are evaluated at $\mathbf{u}$. This formulation lets Beta primitives and the medium MLP jointly explain appearance, attenuation, and backscatter in a unified renderer. In the dynamic mode, frame time is appended to the primitive query, $\mathbf{q}_i(t)=[\mathbf{d}_i,t]$, and to the ambient field through a learned embedding $\gamma(t)$, making the conditioned primitive geometry, opacity, and medium maps time-dependent.

\noindent \textbf{Scattering-aware geometric gradients.}
\label{sec:sgg}
Underwater rasterization couples scene geometry with the medium field, causing projected-center updates to explain backscatter rather than object structure. We therefore introduce Scattering-aware Geometric Gradients (SGG), a medium-aware projected-center preconditioner used during coarse optimization. SGG leaves the forward model unchanged while filtering projected-center gradients, and is coupled with a post-Adam rescaling of 3D center steps.
As summarized in Alg.~\ref{alg:sgg}, the backward pass decomposes each pixel--primitive alpha adjoint into a clear-object branch and a medium branch containing backscatter and attenuation-induced terms. Their signed sums give $\bm{g}^{\mathrm{geo}}_i$ and $\bm{g}^{\mathrm{med}}_i$, while $\bar{\bm{g}}^{\mathrm{geo}}_i$ accumulates component-wise absolute geometry gradients. Using medium uncertainty, SGG preserves the geometry-side direction, recovers unsigned geometry support, and retains a small exact-gradient residual with $\eta=0.15$. A bounded post-Adam multiplier then increases the center displacement of medium-uncertain primitives, gated by their isolated depth-loss gradient to protect depth-consistent geometry. SGG and center-step rescaling are applied for the first $10{,}000$ steps.

\noindent \textbf{Depth regularization for Beta primitives.}
\label{sec:depth_reg}
The feed-forward SfM provides a dense depth prior, which we use to regularize the rendered UWBS on reliable pixels according to
\begin{equation}
\mathcal{L}_{\mathrm{depth}}
=
\sum_{n=1}^{N}
w_n
\log\!\left(
1+\left|\hat{D}_{n}^{-1}-D_{n}^{-1}\right|
\right),
\label{eq:depth_reg}
\end{equation}
where $N$ denotes the number of valid pixels, $\hat{D}_{n}$ is the rendered expected depth, $D_n$ is the SfM depth prior, and $w_n$ is an edge-aware weight. The logarithmic penalty on inverse depth reduces the influence of large absolute-depth errors while preserving near-surface geometric detail.

\begin{algorithm}[t]
\caption{Scattering-aware Geometric Gradients}
\label{alg:sgg}
\begin{algorithmic}[1]
\State \textbf{Input:} Beta-primitive centers $\bm{x}_i$,
cutoff step $\tau_c{=}10{,}000$
\State \textbf{Hyperparameters:} residual $\eta$, confidence/depth policy $f$
\State \textbf{Output:} Updated Beta-primitive centers $\bm{x}_i$
\State \textbf{Current step:} $t$
\State Derive standard projected-center gradients $\{\bm{g}^{0}_i\}$
\If{$t>\tau_c$}
    \State $\bm{x}\gets\mathrm{Adam}(\bm{x},\{\bm{g}^{0}_i\})$
    \State \Return $\bm{x}$
\EndIf
\State Derive per-pixel split gradients
\Statex \hspace{\algorithmicindent}
$\bm{r}^{\mathrm{geo}}_{i,n},\bm{r}^{\mathrm{med}}_{i,n}$
for pixels $n\in\mathcal{N}_i$ covered by primitive $i$.
\State Accumulate signed and unsigned gradients:
\Statex {\scriptsize
\[
\bm{g}^{\mathrm{geo}}_i=\sum_{n\in\mathcal{N}_i}\bm{r}^{\mathrm{geo}}_{i,n},
\quad
\bar{\bm{g}}^{\mathrm{geo}}_i=\sum_{n\in\mathcal{N}_i}|\bm{r}^{\mathrm{geo}}_{i,n}|,
\quad
\bm{g}^{\mathrm{med}}_i=\sum_{n\in\mathcal{N}_i}\bm{r}^{\mathrm{med}}_{i,n}.
\]
}
\For{each Beta primitive $i$ in parallel}
    \State {\footnotesize $u_i^{\mathrm{med}}\gets
    \dfrac{\|\bm{g}^{\mathrm{med}}_i\|_1}
    {\|\bm{g}^{\mathrm{geo}}_i\|_1+\|\bm{g}^{\mathrm{med}}_i\|_1+\varepsilon}$}
    \Comment{medium uncertainty}
    \State {\footnotesize $\kappa_i^{\mathrm{geo}}\gets
    \left[
    \dfrac{\|\bar{\bm{g}}^{\mathrm{geo}}_i\|_1-\|\bm{g}^{\mathrm{geo}}_i\|_1}
    {\|\bar{\bm{g}}^{\mathrm{geo}}_i\|_1+\varepsilon}
    \right]_{0}^{1}$}
    \Comment{signed-sum cancellation}
    \State {\footnotesize $q_i^{\mathrm{geo}}\gets
    (1-u_i^{\mathrm{med}})\kappa_i^{\mathrm{geo}}$}
    \Comment{geometry recovery}
    \State {\footnotesize $\bm{d}^{\mathrm{geo}}_i\gets
    \operatorname{sign}(\bm{g}^{\mathrm{geo}}_i)$}
    \Comment{geometry direction}
    \State {\footnotesize $\bm{g}^{\mathrm{filtered}}_i\gets
    \bm{d}^{\mathrm{geo}}_i\odot
    \big[
    (1-q_i^{\mathrm{geo}})|\bm{g}^{\mathrm{geo}}_i|
    +q_i^{\mathrm{geo}}\bar{\bm{g}}^{\mathrm{geo}}_i
    \big]$}
    \State {\footnotesize $\bm{g}^{\mathrm{sgg}}_i\gets
    \bm{g}^{\mathrm{filtered}}_i+\eta u_i^{\mathrm{med}}\bm{g}^{0}_i$}
    \State {\footnotesize $c_i^{\mathrm{med}}\gets
    f\!\left(
    u_i^{\mathrm{med}};
    \left\|\nabla_{\bm{x}_i}^{\mathrm{iso}}
    \mathcal{L}_{\mathrm{depth}}\right\|_2
    \right)$}
    \Comment{post-Adam center-step multiplier}
\EndFor
\State $\bm{x}^{\mathrm{pre}}\gets\bm{x};\quad
\bm{x}\gets\mathrm{Adam}\big(\bm{x},\{\bm{g}^{\mathrm{sgg}}_i\}\big)$
\State $\bm{x}_i\gets \bm{x}^{\mathrm{pre}}_i+
c_i^{\mathrm{med}}(\bm{x}_i-\bm{x}^{\mathrm{pre}}_i)$
\State \Return $\bm{x}$
\end{algorithmic}
\end{algorithm}

\noindent \textbf{Beta primitive and pose joint optimization.}
Following 3R-GS~\cite{huang20253r}, UWBS jointly optimizes training poses,
primitives, and the medium field using a zero-initialized MLP over per-camera
codes, plus an early MASt3R epipolar term. Test frames are excluded and
handled only by the evaluation protocol.

\noindent \textbf{Training objective.}
UWBS optimizes the standard 3DGS photometric
loss~\cite{kerbl20233d},
\begingroup
\small
\begin{equation}
\mathcal{L}_{\mathrm{3DGS}}
=
(1-\lambda_{\mathrm{ssim}})
\left\|\hat{\mathbf{I}}-\mathbf{I}\right\|_{1}
+
\lambda_{\mathrm{ssim}}
\left(1-\mathrm{SSIM}(\hat{\mathbf{I}},\mathbf{I})\right).
\label{eq:3dgs_loss}
\end{equation}
\endgroup

We also use the opacity and scale losses of
MCMC-GS~\cite{kheradmand20243d},
\begin{equation}
\mathcal{L}_{\mathrm{MCMC}}
=
\lambda_{\mathrm{opa}}\mathcal{L}_{\mathrm{opa}}
+
\lambda_{\mathrm{scale}}\mathcal{L}_{\mathrm{scale}} .
\label{eq:mcmc_reg}
\end{equation}
In addition, we regularize the medium field with smoothness,
\begin{equation}
\mathcal{L}_{\mathrm{smooth}}^{\mathrm{med}}
=
\frac{1}{N_x}\sum_{n=1}^{N_x}
\left|\Delta_x \mathbf{M}_{n}\right|
+
\frac{1}{N_y}\sum_{n=1}^{N_y}
\left|\Delta_y \mathbf{M}_{n}\right|,
\label{eq:medium_smooth}
\end{equation}
where $N_x$ and $N_y$ are the numbers of valid horizontal and vertical finite
differences. To prevent oversized Beta primitives from acting as veil proxies,
we use a projected-radius loss,
\begin{equation}
\mathcal{L}_{\mathrm{rad}}
=
\operatorname*{avg}_{i:\,r_i^{\mathrm{uncap}}>256\mathrm{px}}
\left(\frac{1}{3}\left\|\mathbf{s}_i\right\|_{1}\right),
\label{eq:radius_reg}
\end{equation}
where $r_i^{\mathrm{uncap}}$ is the uncapped projected radius and
$\mathbf{s}_i$ is the 3D spatial scale of primitive $i$.

Finally, our total learning objective of UWBS, together with the depth loss
defined in \eqnref{eq:depth_reg}, is
\begin{equation}
\begin{aligned}
\mathcal{L}_{\mathrm{total}}
&=
\mathcal{L}_{\mathrm{3DGS}}
+
\mathcal{L}_{\mathrm{MCMC}}
+
\lambda_{\mathrm{depth}}\mathcal{L}_{\mathrm{depth}}
\\
&\quad
+
\lambda_{\mathrm{smooth}}^{\mathrm{med}}\mathcal{L}_{\mathrm{smooth}}^{\mathrm{med}}
+
\lambda_{\mathrm{rad}}\mathcal{L}_{\mathrm{rad}} .
\end{aligned}
\label{eq:uwbs_total}
\end{equation}
We use $\lambda_{\mathrm{ssim}}=0.2$,
$\lambda_{\mathrm{smooth}}^{\mathrm{med}}=1.0$,
$\lambda_{\mathrm{opa}}=0.01$, $\lambda_{\mathrm{scale}}=0.01$,
$\lambda_{\mathrm{depth}}=0.1$, and $\lambda_{\mathrm{rad}}=0.1$.

\begin{figure*}[t]
\centering
\scriptsize
\setlength{\tabcolsep}{1pt}
\renewcommand{\arraystretch}{0}
{\normalsize
\resizebox{0.85\textwidth}{!}{%
\begin{tabular}{@{}M{1em} M{0.245\textwidth} M{0.245\textwidth} M{0.245\textwidth} M{0.245\textwidth}@{}}

& Cyan & Murky & Outcrop & Caustic \\[10pt]

\rotatebox{90}{COLMAP} &
\includegraphics[width=0.245\textwidth,height=3cm,keepaspectratio]{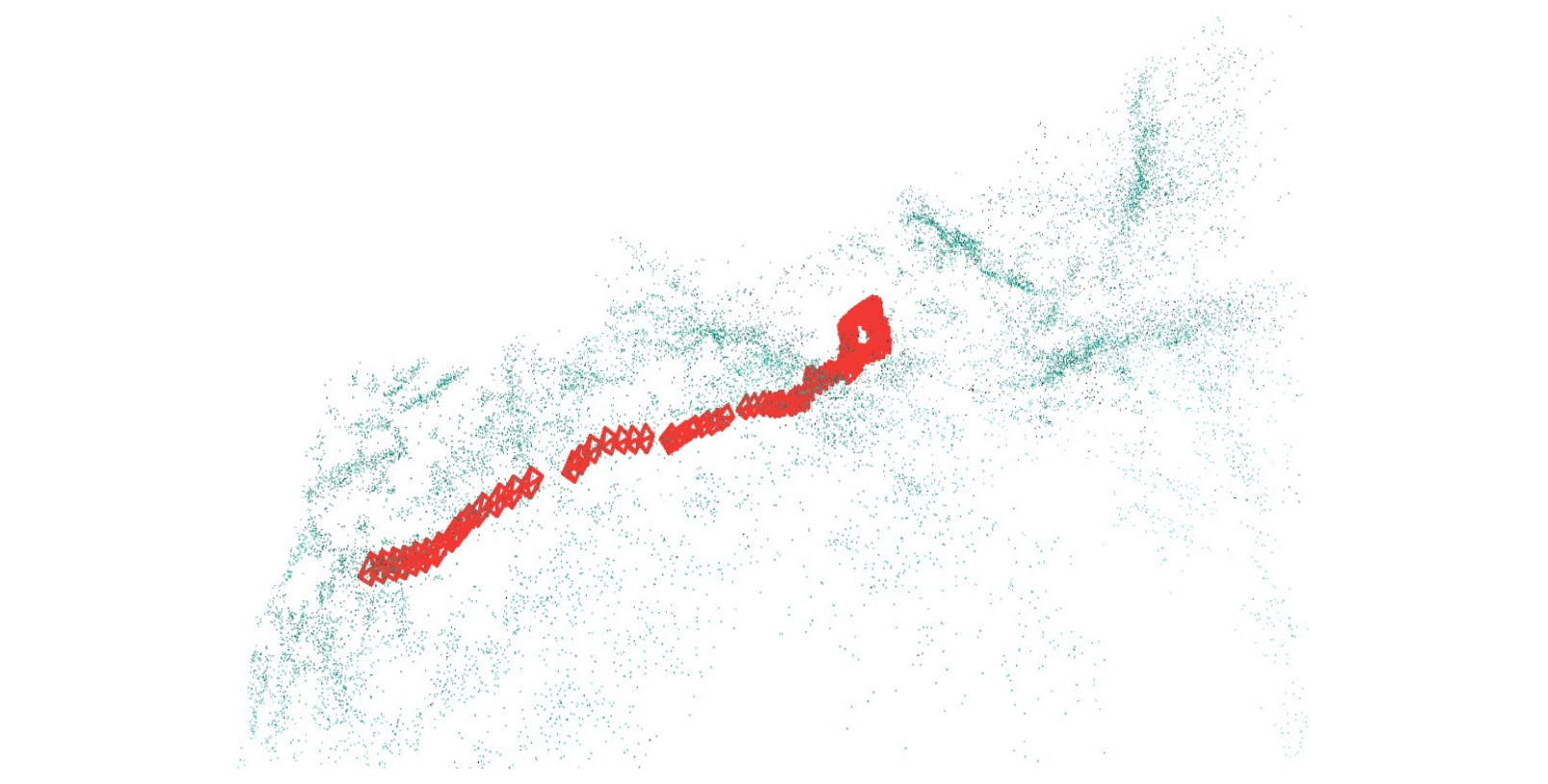} &
\includegraphics[width=0.245\textwidth,height=3cm,keepaspectratio]{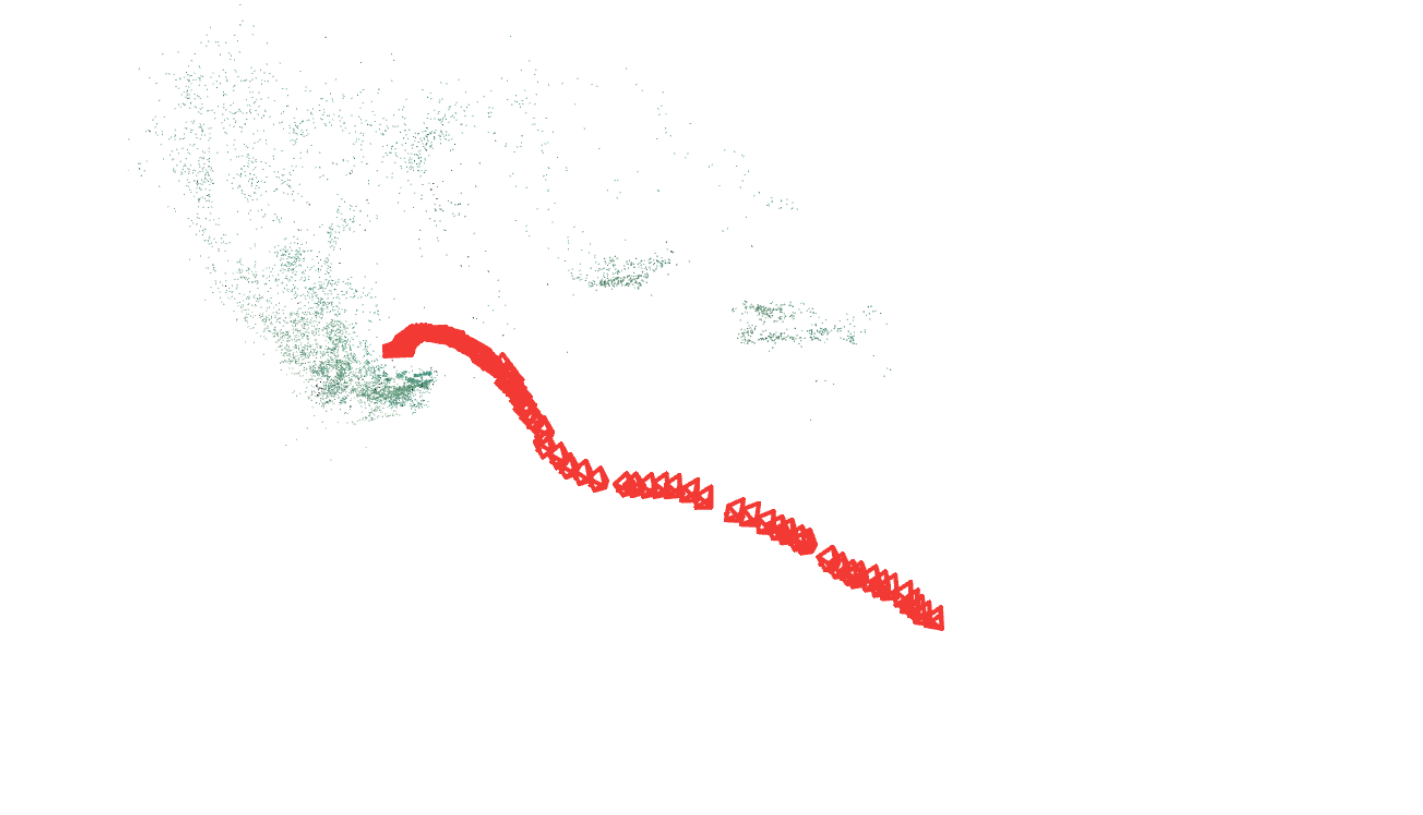} &
\includegraphics[width=0.245\textwidth,height=3cm,keepaspectratio]{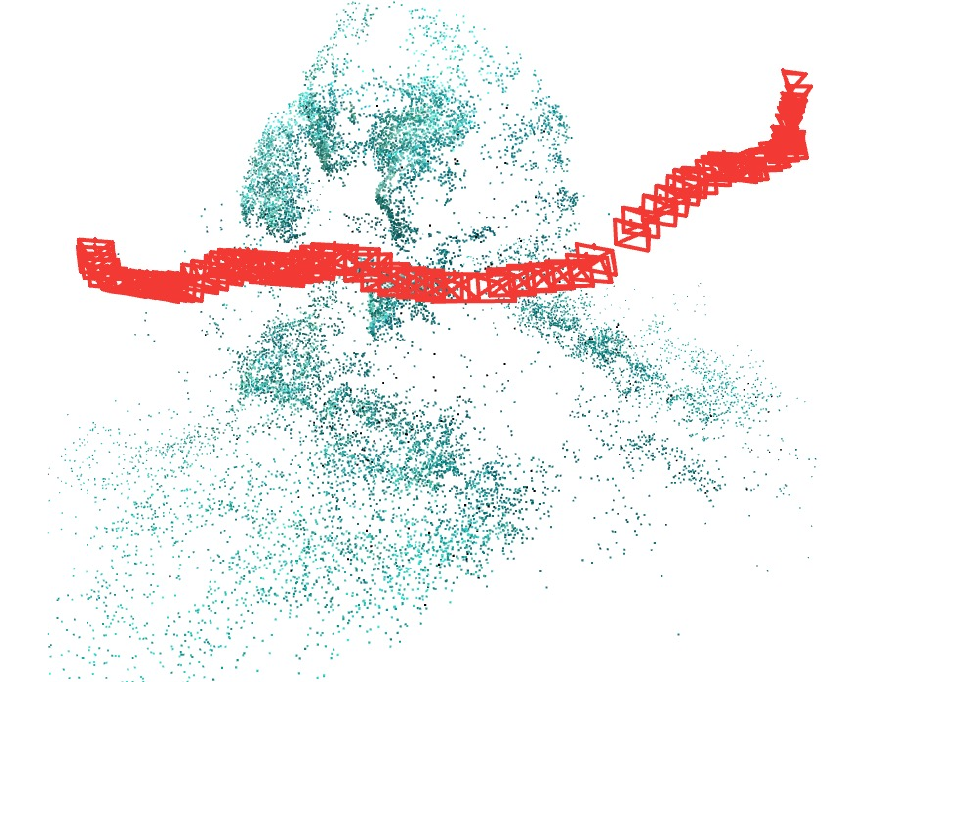} &
\includegraphics[width=0.245\textwidth,height=3cm,keepaspectratio]{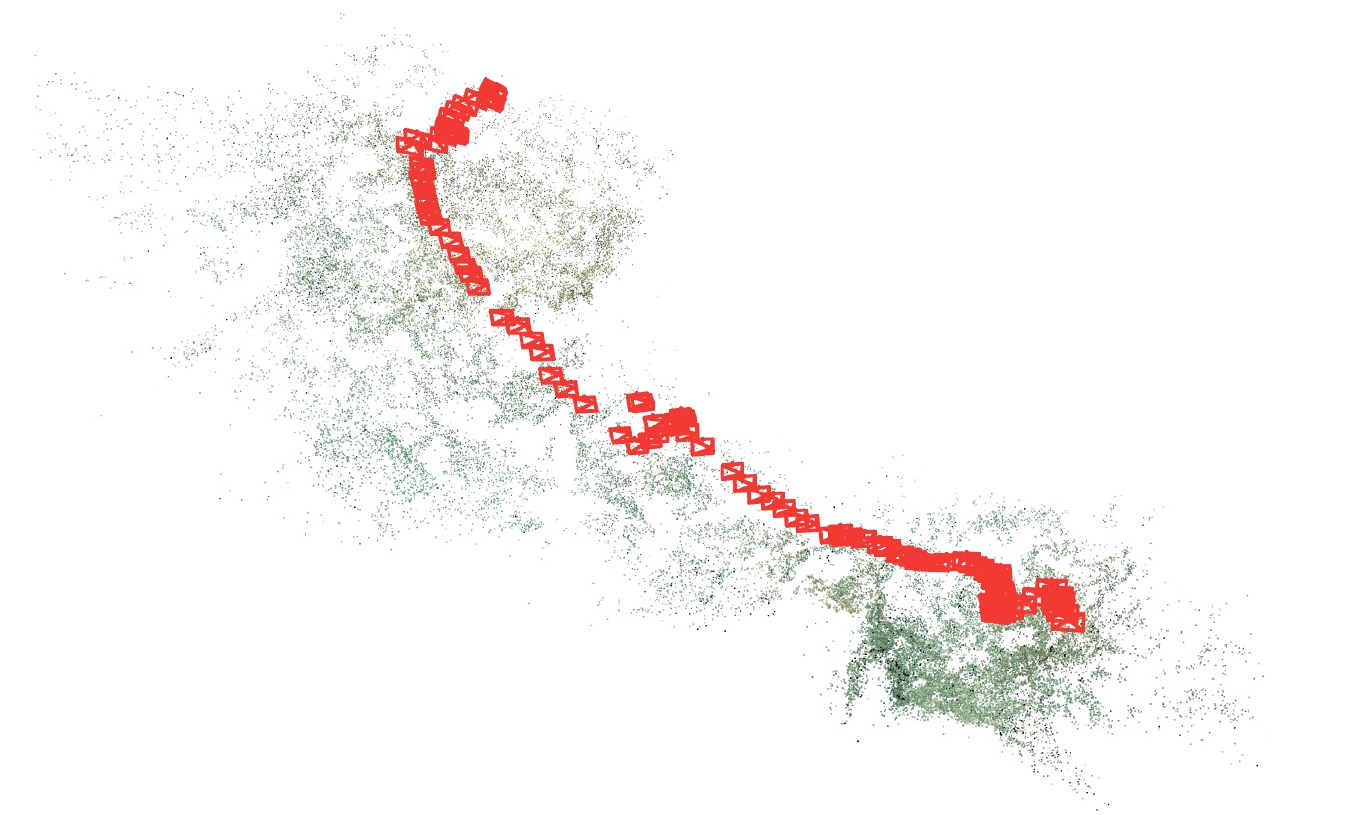} \\[-12pt]

\rotatebox{90}{GLUEMAP} &
\includegraphics[width=0.245\textwidth,height=3cm,keepaspectratio]{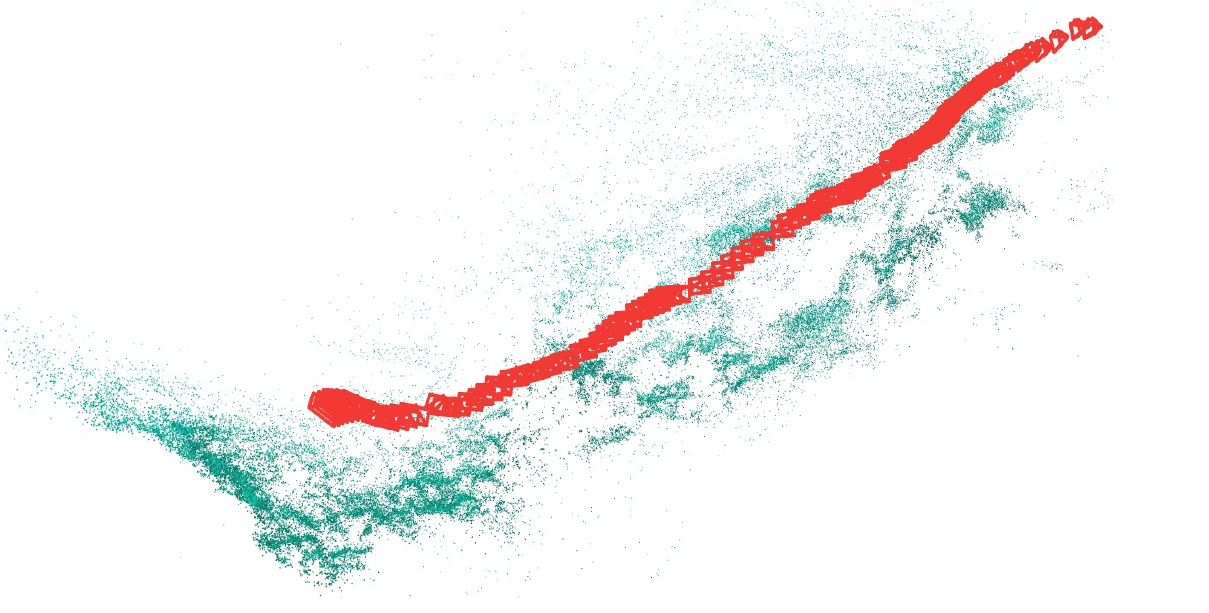} &
\includegraphics[width=0.245\textwidth,height=3cm,keepaspectratio]{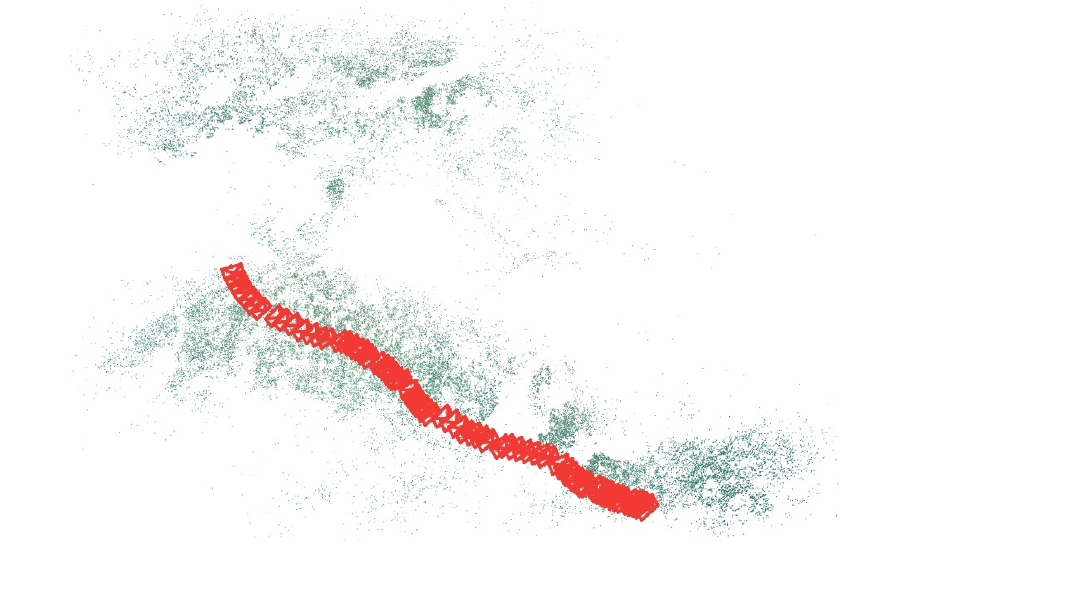} &
\includegraphics[width=0.245\textwidth,height=3cm,keepaspectratio]{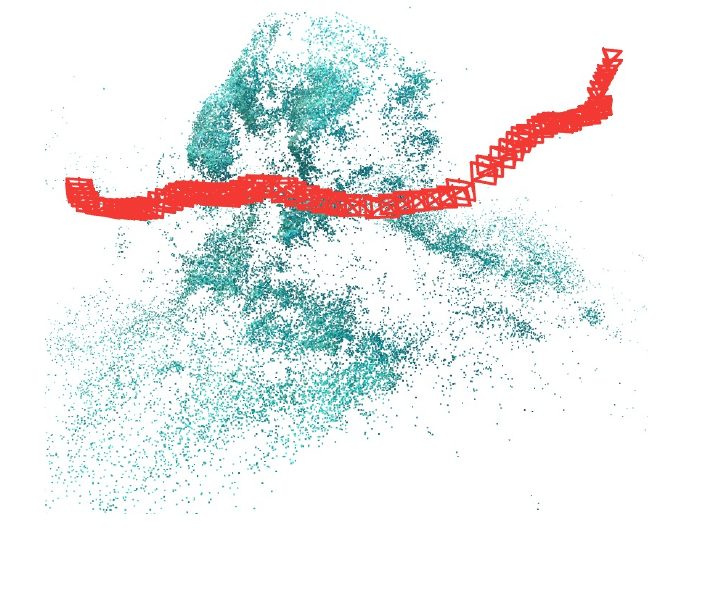} &
\includegraphics[width=0.245\textwidth,height=3cm,keepaspectratio]{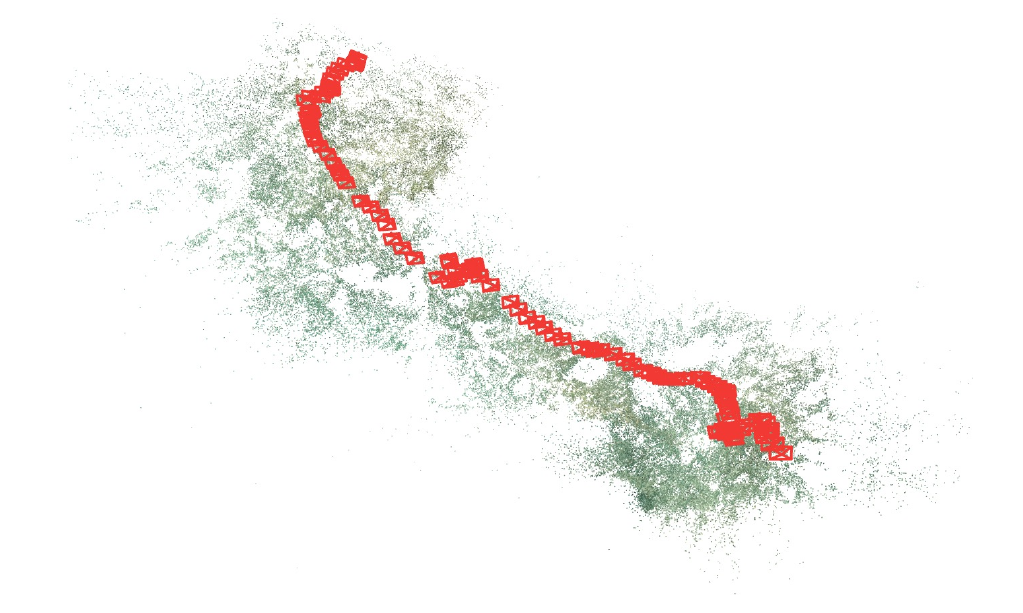} \\[-12pt]

\rotatebox{90}{MASt3R-SfM} &
\includegraphics[width=0.245\textwidth,height=3cm,keepaspectratio]{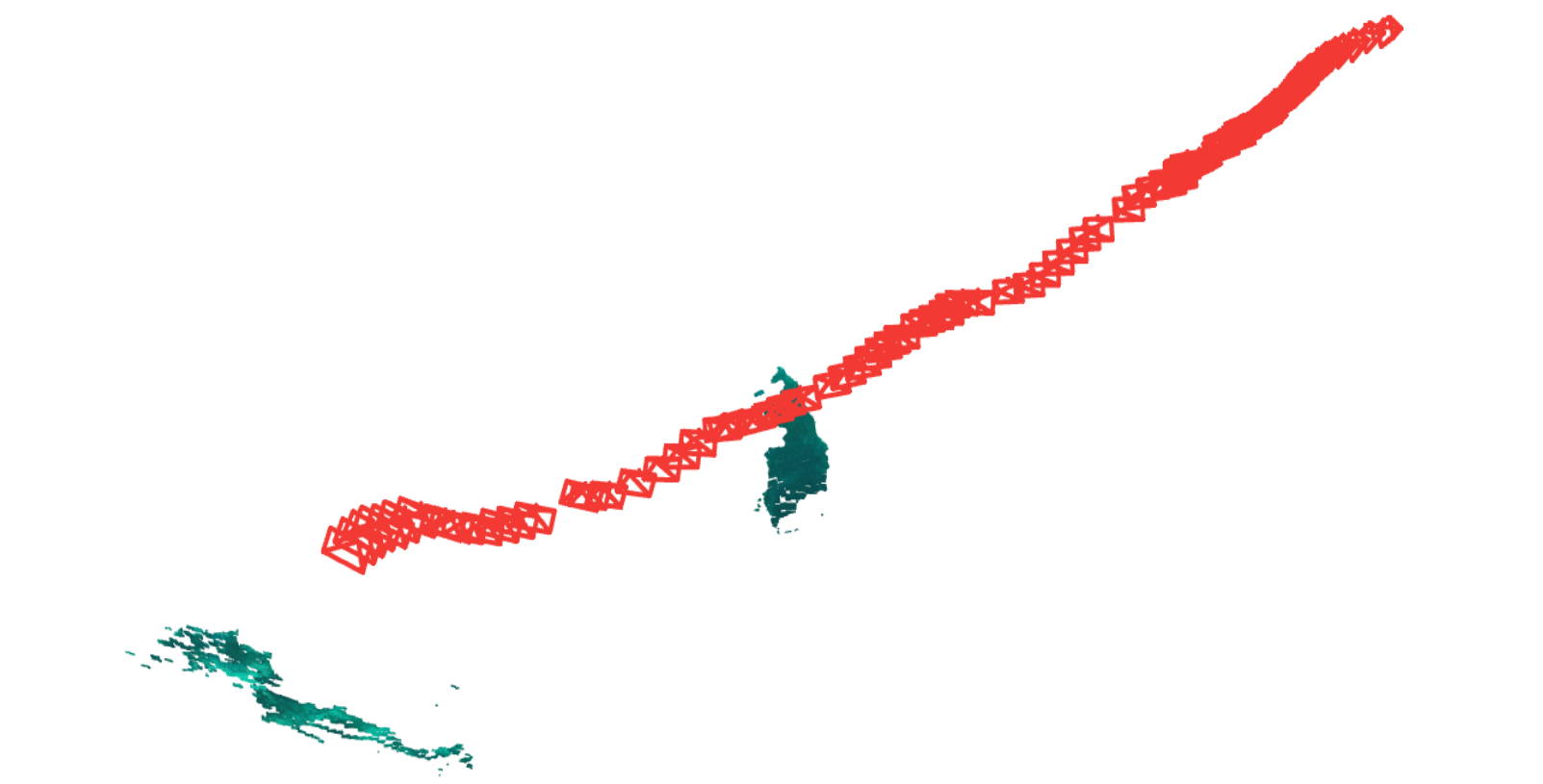} &
\includegraphics[width=0.245\textwidth,height=3cm,keepaspectratio]{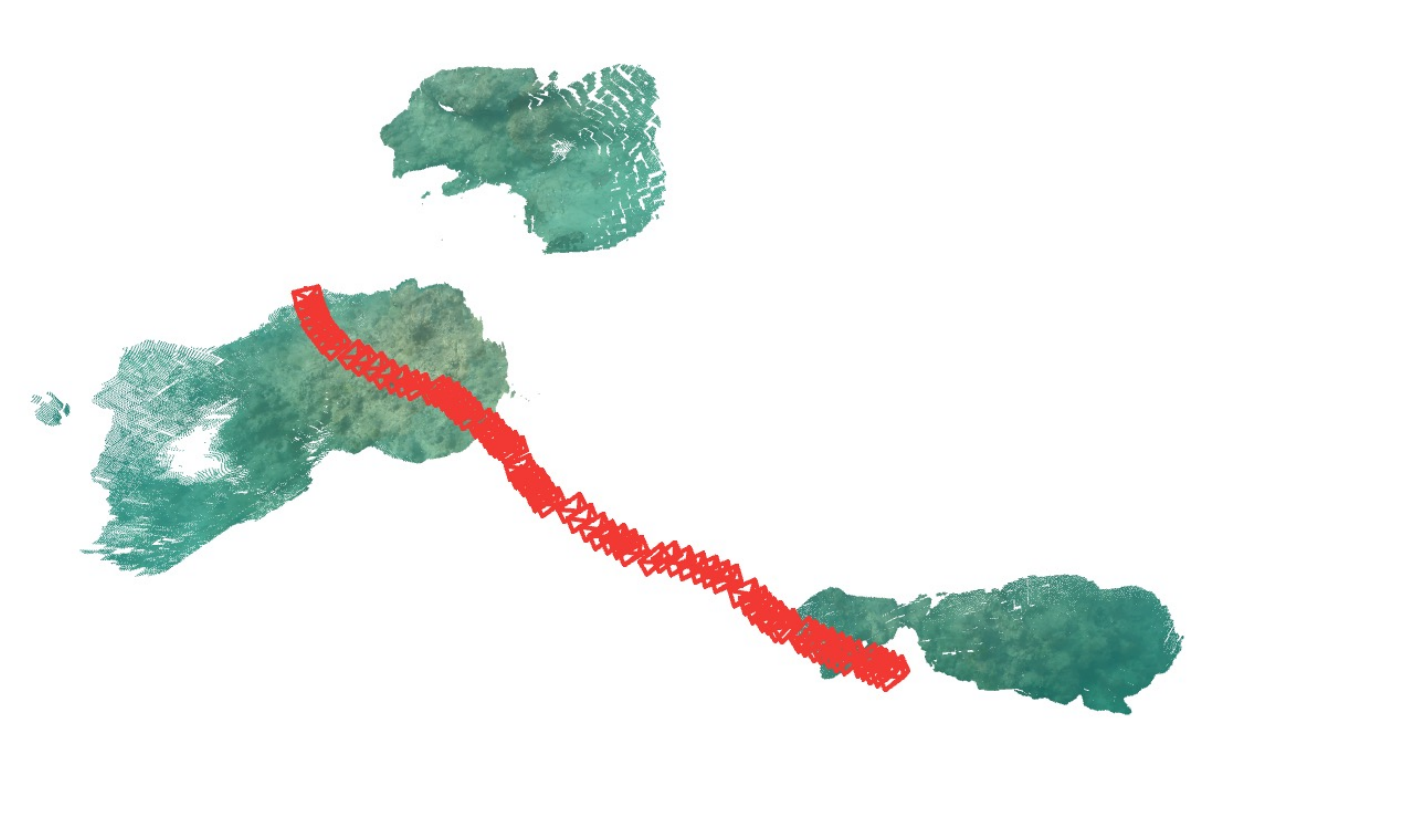} &
\includegraphics[width=0.245\textwidth,height=3cm,keepaspectratio]{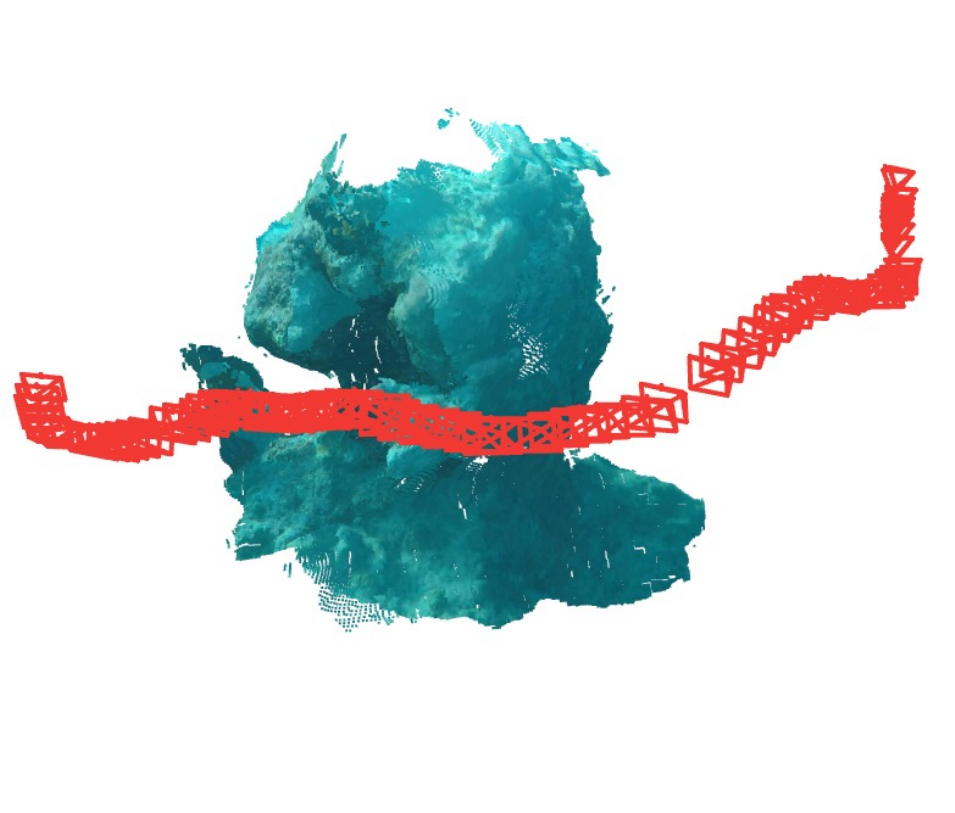} &
\includegraphics[width=0.245\textwidth,height=3cm,keepaspectratio]{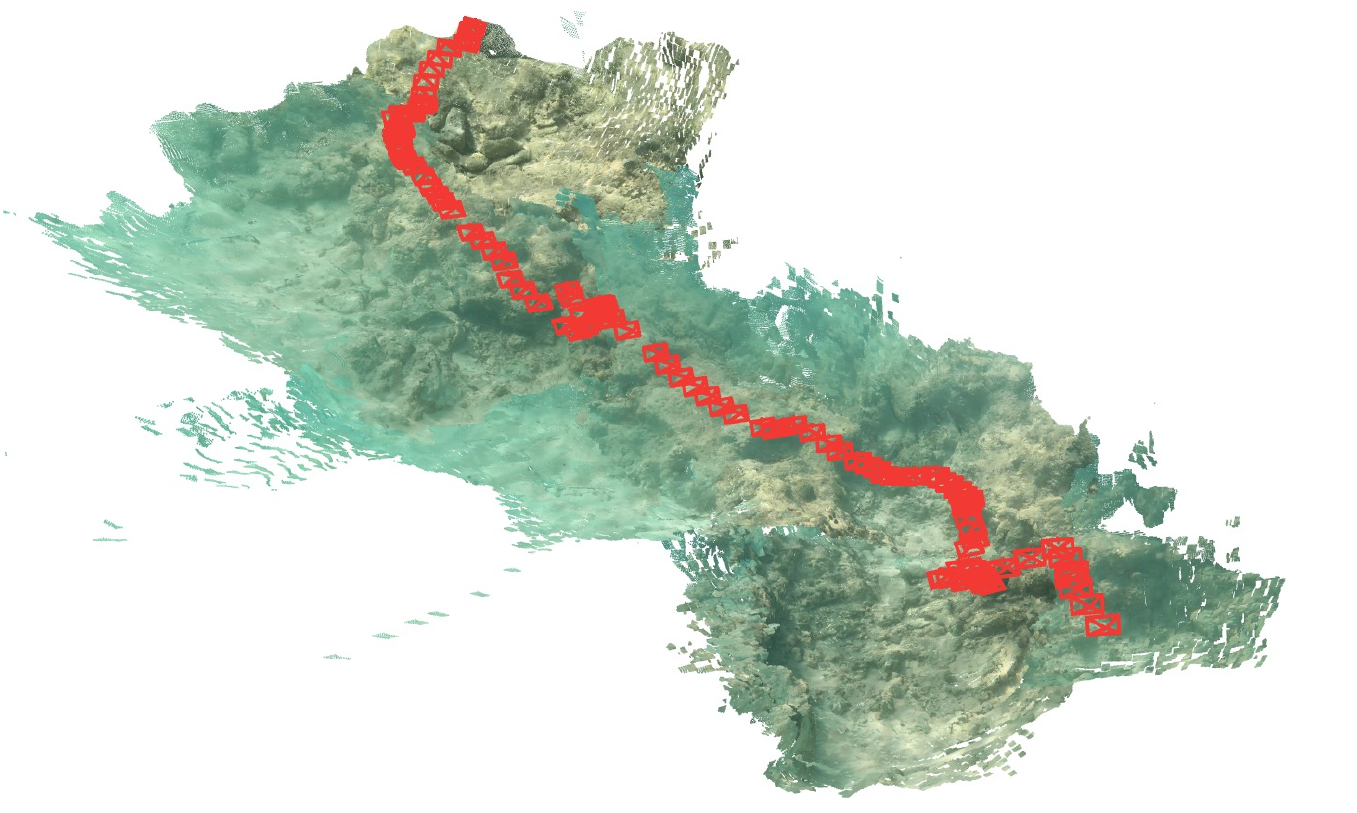} \\[-12pt]

\rotatebox{90}{\textbf{Swimm3R}} &
\includegraphics[width=0.245\textwidth,height=3cm,keepaspectratio]{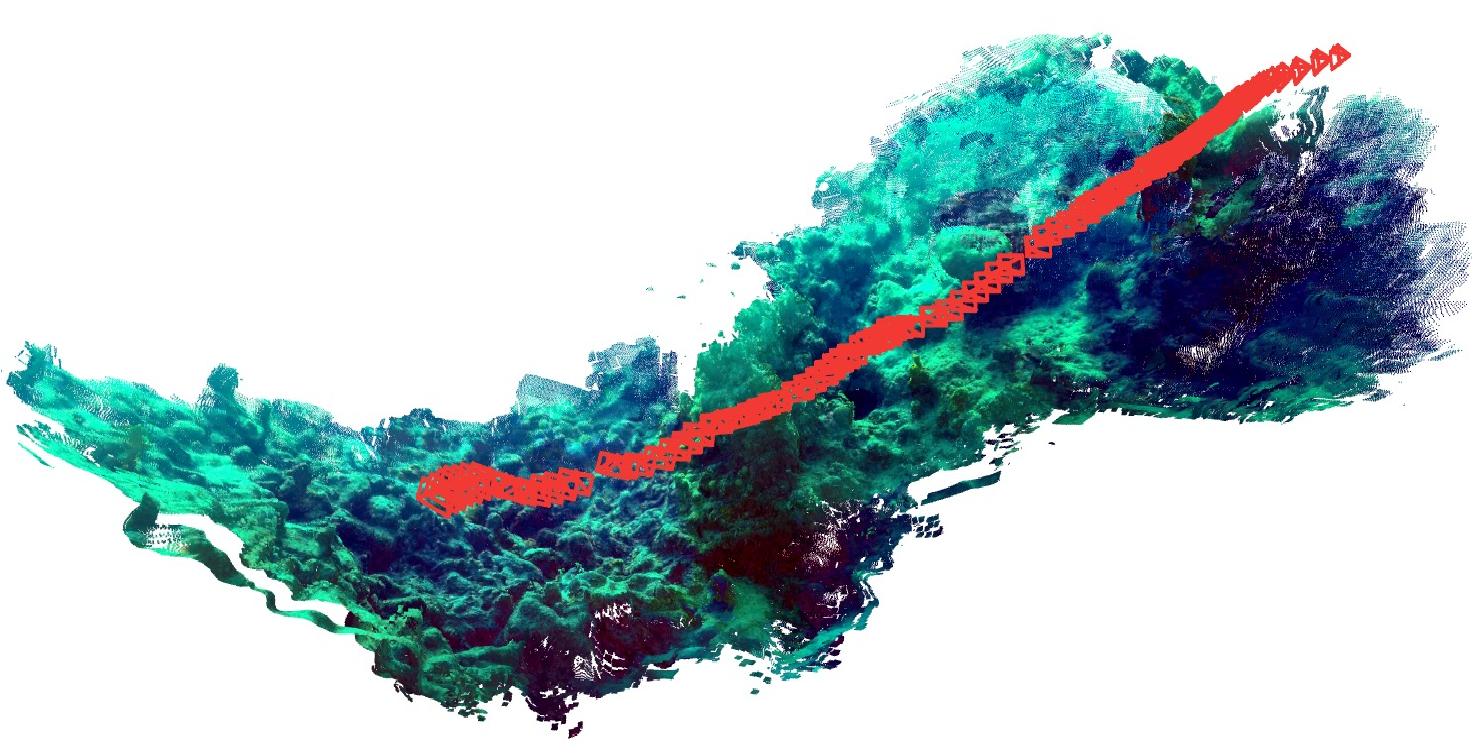} &
\includegraphics[width=0.245\textwidth,height=3cm,keepaspectratio]{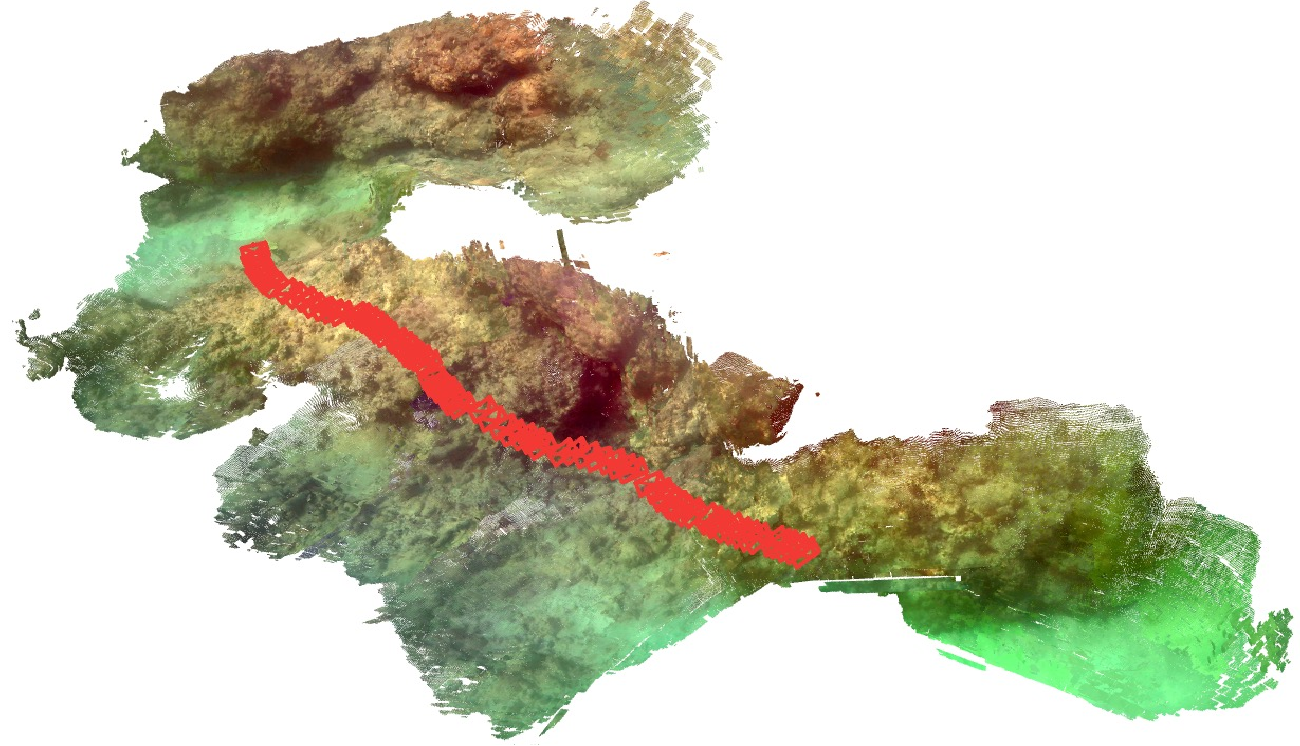} &
\includegraphics[width=0.245\textwidth,height=3cm,keepaspectratio]{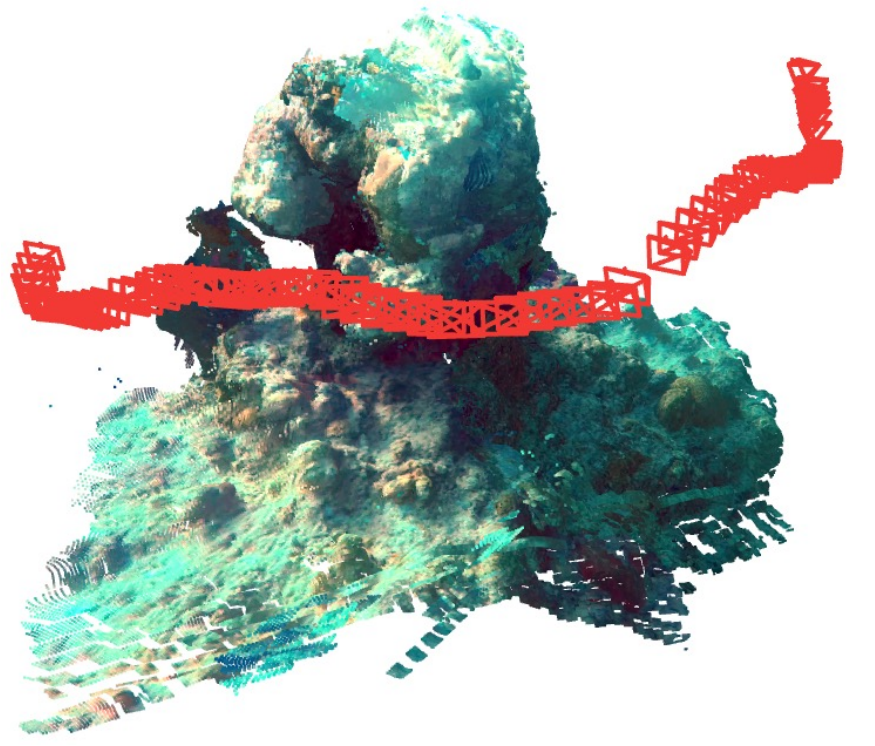} &
\includegraphics[width=0.245\textwidth,height=3cm,keepaspectratio]{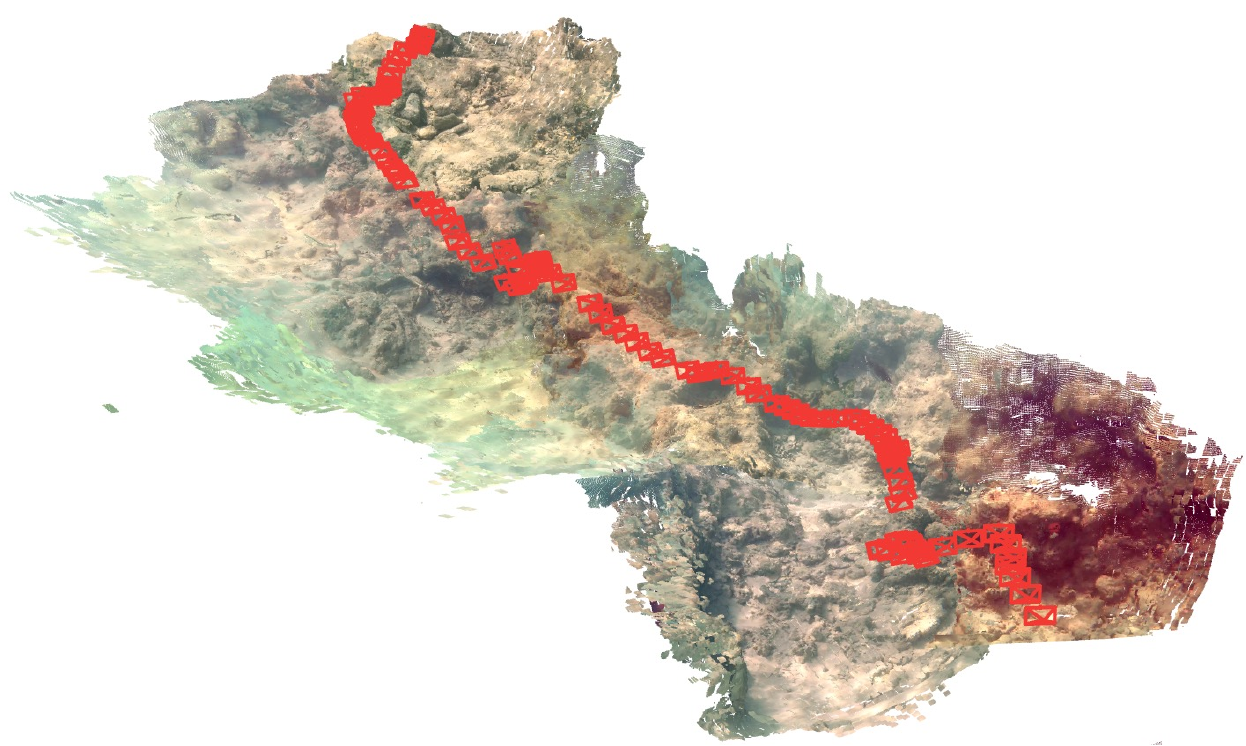} \\
\end{tabular}%
}
}
\caption{Qualitative comparison of underwater 3D reconstruction. Swimm3R yields more continuous and geometrically coherent scene structures than prior SfM methods.}
\label{fig:point_map_comparison}
\end{figure*}

\section{Experiments}
\subsection{Datasets}
For medium-aware SfM training, we use the FLSea~\cite{randall2023flsea} and Sea-Thru~\cite{akkaynak2019sea} datasets.
For the train-test split, sequences are separated at the sequence level, with \texttt{Horse\_Canyon}, \texttt{Tiny\_Canyon}, and \texttt{Red\_Sea\_Sub\_Pier} from the FLSea dataset designated as test sequences.
For underwater 3D mapping and visual localization, we use our Barbados dataset, subsample each 30 fps video to 5 fps, and reserve every $8^{th}$ frame as test and query frames. All SfM and splatting optimization use only the remaining training frames.

\subsection{Metrics}
\label{sec:metric}
On FLSea, we evaluate depth with AbsRel, $\delta<1.25$, and $\text{RMSE}_{\log}$, and camera pose with the scale-free metrics RRA@15, RTA@15, and mAA(30), following MASt3R-SfM~\cite{duisterhof2025mast3r}, taking COLMAP on the corresponding clean sequences as reference poses for scale-free pose evaluation only. On Barbados, where ground-truth geometry is unavailable, we instead report SfM map quality as the frame registration rate, the mean symmetric epipolar error, the $1$-px inlier ratio of retained two-view correspondences, and the number of reconstructed 3D points, with both correspondence metrics in pixels under each method's own poses. Restored appearance is scored by the no-reference metrics UIQM~\cite{panetta2015human}, UCIQE~\cite{yang2015underwater}, and UnderwaterRanker~\cite{guo2023uranker}, and test-frame rendering by PSNR, SSIM, and LPIPS~\cite{zhang2018unreasonable}, following 3DGS~\cite{kerbl20233d}.

For visual localization we follow GS-CPR~\cite{hao2025camera}: the top-$5$ training frames retrieved by MASt3R encoder cosine similarity are re-matched into a coarse pose, from which we render underwater RGB and depth, match the render to the query, and unproject the matches for PnP-RANSAC~\cite{lepetit2009ep}. All renderers share the same Swimm3R point clouds and poses, isolating the renderer. Since COLMAP fails to register the full Barbados sequences, GLUEMAP~\cite{pan2026global} poses serve as the reference.

\begin{table}[t]
\centering
\caption{Depth and pose estimation on FLSea test sequences.}
\label{tab:swimm3r_flsea}
\resizebox{\linewidth}{!}{%
{\Large
\begin{tabular}{l|ccc|ccc}
 & \multicolumn{3}{c|}{\textit{Depth}}
 & \multicolumn{3}{c}{\textit{Pose}} \\
Method
& AbsRel$\downarrow$ & $\delta\!<\!1.25\uparrow$ & RMSE$_{\log}\downarrow$
& RRA@15$\uparrow$ & RTA@15$\uparrow$ & mAA(30)$\uparrow$ \\
\hline
Spann3R~\cite{wang20253d}
& 0.2964 & 73.50 & 0.2908
& 20.25 & 8.67 & 3.45 \\
VGGT~\cite{wang2025vggt}
& 0.2539 & 89.41 & 0.2071
& 56.01 & 29.44 & 23.02 \\
MASt3R-SfM~\cite{duisterhof2025mast3r}
& \ranktwo{0.2247} & \ranktwo{93.66} & \ranktwo{0.2006}
& \ranktwo{77.47} & \ranktwo{86.81} & \ranktwo{66.83} \\
Dark3R~\cite{guo2026dark3r}
& \rankthree{0.2288} & \rankthree{93.07} & \rankthree{0.2049}
& \rankthree{63.89} & \rankthree{81.47} & \rankthree{56.37} \\
\textbf{Swimm3R (Ours)}
& \rankone{0.2132} & \rankone{95.11} & \rankone{0.1915}
& \rankone{79.19} & \rankone{88.43} & \rankone{69.67} \\
\hline
\end{tabular}
}
}%
{\footnotesize \quad \ranklegend}
\end{table}

\subsection{Implementation details}

\noindent \textbf{Medium-aware Structure-from-Motion.}
We initialize the backbone from the publicly released MASt3R~\cite{leroy2024grounding} checkpoint and zero-initialize all LoRA~\cite{hu2022lora} adapters. For the LoRA setup, we use rank $16$ for the patch embedding and projection layers, and rank $8$ for the decoder cross-attention projections. We optimize with AdamW using a learning rate of $5{\times}10^{-4}$, weight decay of $10^{-4}$, and train for $20$ epochs with batch size $4$. Each training run is conducted on a single RTX A6000 Pro GPU and trained for approximately $4.5$ hours.

\noindent \textbf{Underwater Beta Splatting.}
We initialize UWBS from the point cloud produced by our medium-aware SfM through voxelized sampling, yielding approximately $400$K spatially distributed primitives. UWBS employs MCMC-based densification~\cite{kheradmand20243d} and caps the map at $800$K primitives. Each scene is trained for $20$K iterations, with both our SGG module and the densification process active during the first half. All remaining learning rates follow those of our baseline Universal Beta Splatting (UBS)~\cite{liu2025universal}.

\begin{table}[t]
\centering
\caption{SfM 3D map quality on the Barbados dataset.}
\label{tab:sfm_metrics}
\resizebox{0.4\textwidth}{!}{%
{\large
\begin{tabular}{l|ccc @{\hskip 1.2em} c}
Method & Reg.\,(\%)\,$\uparrow$ & EpiErr\,$\downarrow$ & Inlier@1px\,(\%)\,$\uparrow$ & \#Points \\
\hline
COLMAP     & \ranktwo{87.7} & 3.577 & 61.6 & 21.8k \\
GLUEMAP    & \rankone{100.0} & \rankthree{0.720} & \rankthree{69.2} & 59.6k \\
MASt3R-SfM & \rankone{100.0} & \ranktwo{0.688} & \ranktwo{70.5} & 1.01M \\
\textbf{Swimm3R (Ours)}
           & \rankone{100.0} & \rankone{0.667} & \rankone{71.5} & 2.19M \\
\hline
\end{tabular}
}
}%
\end{table}

\begin{figure*}[t]
\centering
\scriptsize
\setlength{\tabcolsep}{1.5pt}
\setlength{\arrayrulewidth}{0.6pt}%
\setlength{\dashlinedash}{2pt}%
\setlength{\dashlinegap}{1.5pt}%
\resizebox{0.9\textwidth}{!}{%
\begin{tabular}{@{}m{0.8em} m{0.15\textwidth} m{0.15\textwidth} : m{0.15\textwidth} m{0.15\textwidth} : m{0.15\textwidth} m{0.15\textwidth}@{}}
 & \multicolumn{2}{c:}{SeaSplat} & \multicolumn{2}{c:}{WaterSplatting} & \multicolumn{2}{c}{\textbf{UWBS}} \\
\smash{\rotatebox[origin=c]{90}{GLUEMAP}}
 & \shortstack{\includegraphics[width=0.15\textwidth]{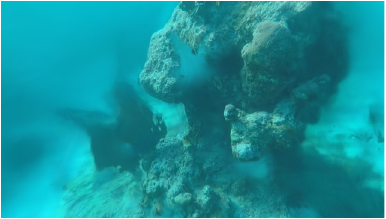}
}
 & \shortstack{\includegraphics[width=0.15\textwidth]{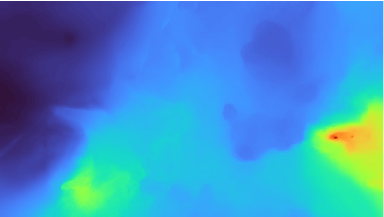}
}
 & \shortstack{\includegraphics[width=0.15\textwidth]{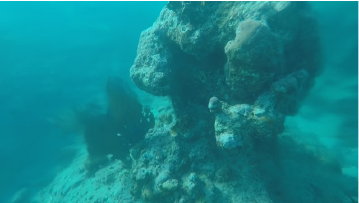}
}
 & \shortstack{\includegraphics[width=0.15\textwidth]{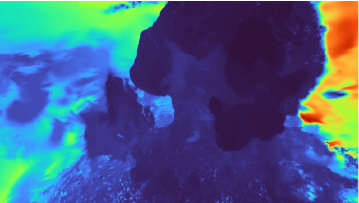}
}
 & \shortstack{\includegraphics[width=0.15\textwidth]{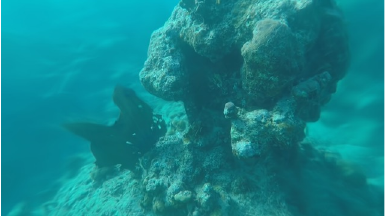}
}
 & \shortstack{\includegraphics[width=0.15\textwidth]{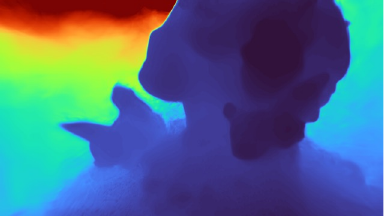}
} \\
\smash{\rotatebox[origin=c]{90}{MASt3R-SfM}}
 & \shortstack{\includegraphics[width=0.15\textwidth]{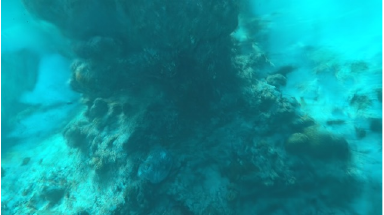}
}
 & \shortstack{\includegraphics[width=0.15\textwidth]{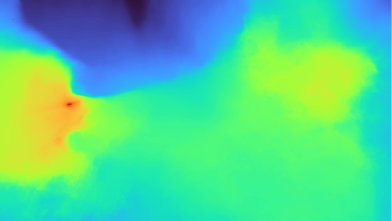}
}
 & \shortstack{\includegraphics[width=0.15\textwidth]{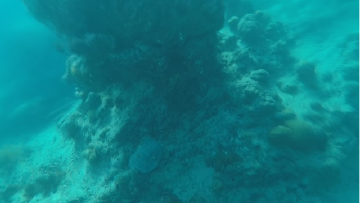}
}
 & \shortstack{\includegraphics[width=0.15\textwidth]{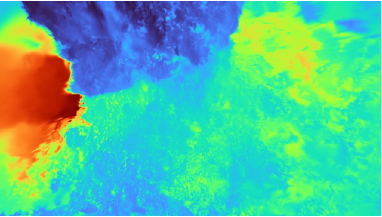}
}
 & \shortstack{\includegraphics[width=0.15\textwidth]{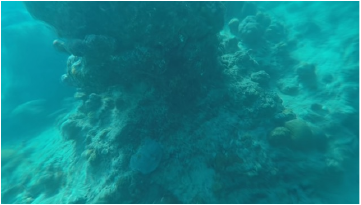}
}
 & \shortstack{\includegraphics[width=0.15\textwidth]{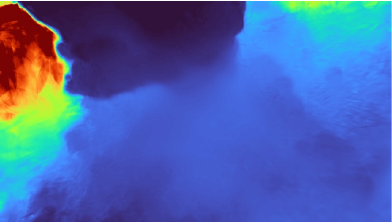}
} \\
\smash{\rotatebox[origin=c]{90}{\textbf{Swimm3R}}}
 & \shortstack{\includegraphics[width=0.15\textwidth]{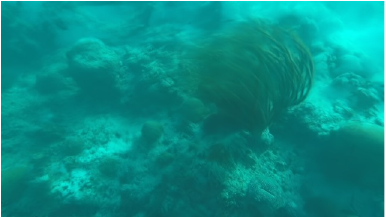}
}
 & \shortstack{\includegraphics[width=0.15\textwidth]{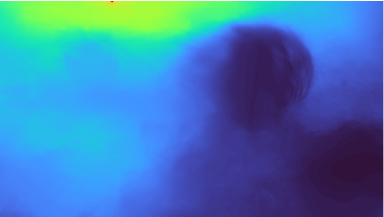}
}
 & \shortstack{\includegraphics[width=0.15\textwidth]{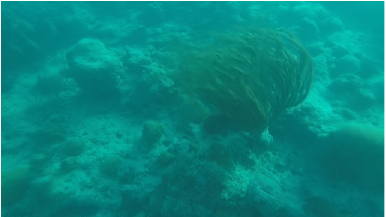}
}
 & \shortstack{\includegraphics[width=0.15\textwidth]{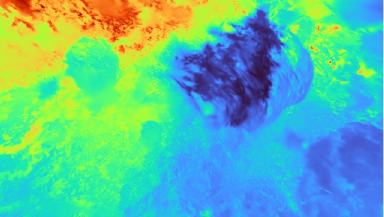}
}
 & \shortstack{\includegraphics[width=0.15\textwidth]{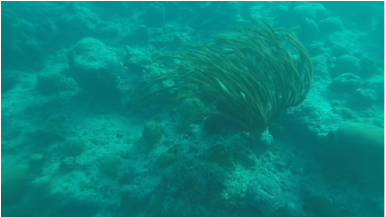}
}
 & \shortstack{\includegraphics[width=0.15\textwidth]{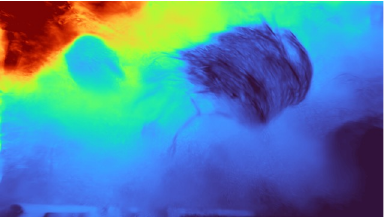}
} \\
\end{tabular}%
}
\caption{Qualitative comparison of RGB and depth across SfM initializations (rows) and renderers (columns). UWBS preserves sharper scene structure and more consistent medium boundaries than prior methods, qualitatively suggesting more coherent appearance and depth under underwater scattering.}
\label{fig:render_results}
\end{figure*}

\begin{table*}[t]
\centering
\caption{Quantitative comparison of test-frame rendering quality on the Barbados dataset across feed-forward SfM initializations and renderers.}
\label{tab:quantitative_comparison}
\setlength{\tabcolsep}{3pt}
\renewcommand{\arraystretch}{1.15}
\setlength{\dashlinedash}{2pt}\setlength{\dashlinegap}{1.5pt}
\resizebox{\textwidth}{!}{%
{\scriptsize
\begin{tabular}{l@{\hspace{3pt}}|@{\hspace{3pt}}l|ccc|ccc|ccc|ccc|ccc}
\multicolumn{2}{c|}{Method} & \multicolumn{3}{c|}{Cyan} & \multicolumn{3}{c|}{Murky} & \multicolumn{3}{c|}{Outcrop} & \multicolumn{3}{c|}{Caustic} & \multicolumn{3}{c}{\textbf{Average}} \\
\cline{1-2}\cline{3-17}
\multicolumn{1}{c}{SfM} & \multicolumn{1}{c|}{Renderer} & PSNR$\uparrow$ & SSIM$\uparrow$ & LPIPS$\downarrow$ & PSNR$\uparrow$ & SSIM$\uparrow$ & LPIPS$\downarrow$ & PSNR$\uparrow$ & SSIM$\uparrow$ & LPIPS$\downarrow$ & PSNR$\uparrow$ & SSIM$\uparrow$ & LPIPS$\downarrow$ & PSNR$\uparrow$ & SSIM$\uparrow$ & LPIPS$\downarrow$ \\
\hline
\multirow{5}{*}{GLUEMAP}
 & 3R-GS & 23.55 & 0.775 & 0.619 & 25.65 & 0.902 & 0.404 & 22.91 & 0.785 & 0.582 & 21.47 & 0.735 & 0.692 & 23.40 & 0.799 & 0.574 \\
 & SeaSplat & 22.98 & 0.777 & 0.423 & 25.73 & 0.897 & 0.351 & 19.91 & 0.799 & 0.379 & 23.61 & \ranktwo{0.779} & \rankthree{0.349} & 23.06 & 0.813 & 0.376 \\
 & WaterSplatting & \rankthree{30.29} & \rankthree{0.818} & \rankthree{0.291} & \rankthree{32.50} & \rankthree{0.924} & \rankthree{0.265} & \rankthree{30.18} & \rankthree{0.844} & \rankone{0.260} & \rankthree{24.82} & \rankthree{0.776} & \ranktwo{0.325} & \rankthree{29.45} & \rankthree{0.841} & \rankthree{0.285} \\
\cdashline{2-17}
 & \textbf{Ours} (Static) & \rankone{32.67} & \rankone{0.855} & \rankone{0.258} & \ranktwo{33.46} & \rankone{0.936} & \rankone{0.255} & \rankone{32.02} & \rankone{0.872} & \ranktwo{0.261} & \ranktwo{25.05} & \rankone{0.816} & 0.337 & \ranktwo{30.80} & \rankone{0.870} & \ranktwo{0.277} \\
 & \textbf{Ours} (Dynamic) & \ranktwo{32.61} & \ranktwo{0.854} & \ranktwo{0.260} & \rankone{33.62} & \ranktwo{0.936} & \ranktwo{0.256} & \ranktwo{31.98} & \ranktwo{0.871} & \rankthree{0.261} & \rankone{25.32} & \rankone{0.816} & \rankone{0.325} & \rankone{30.88} & \ranktwo{0.869} & \rankone{0.276} \\
\hline
\multirow{5}{*}{MASt3R-SfM}
 & 3R-GS & 26.08 & \rankthree{0.801} & 0.643 & 29.75 & 0.909 & 0.402 & \rankthree{31.28} & \rankthree{0.856} & \rankone{0.203} & \ranktwo{25.12} & \rankthree{0.796} & \rankone{0.270} & 28.06 & \rankthree{0.841} & 0.380 \\
 & SeaSplat & 23.10 & 0.773 & 0.486 & 25.93 & 0.897 & 0.360 & 20.27 & 0.790 & 0.341 & 22.71 & 0.737 & 0.382 & 23.00 & 0.799 & 0.392 \\
 & WaterSplatting & \rankthree{26.53} & 0.790 & \rankthree{0.465} & \rankthree{32.32} & \rankthree{0.923} & \rankthree{0.282} & 30.77 & 0.851 & 0.263 & 22.99 & 0.740 & 0.379 & \rankthree{28.15} & 0.826 & \rankthree{0.347} \\
\cdashline{2-17}
 & \textbf{Ours} (Static) & \ranktwo{32.54} & \ranktwo{0.843} & \ranktwo{0.340} & \ranktwo{32.88} & \ranktwo{0.931} & \rankone{0.237} & \rankone{32.37} & \rankone{0.876} & \ranktwo{0.218} & \rankthree{24.88} & \ranktwo{0.813} & \rankthree{0.319} & \ranktwo{30.67} & \ranktwo{0.865} & \ranktwo{0.278} \\
 & \textbf{Ours} (Dynamic) & \rankone{32.67} & \rankone{0.844} & \rankone{0.324} & \rankone{33.38} & \rankone{0.932} & \ranktwo{0.240} & \ranktwo{32.26} & \ranktwo{0.875} & \rankthree{0.219} & \rankone{25.35} & \rankone{0.815} & \ranktwo{0.314} & \rankone{30.92} & \rankone{0.867} & \rankone{0.274} \\
\hline
\multirow{5}{*}{\makecell[l]{\textbf{Swimm3R}\\\textbf{(Ours)}}}
 & 3R-GS & \rankthree{31.04} & \rankthree{0.825} & \rankthree{0.239} & 29.55 & 0.909 & 0.400 & \rankthree{31.16} & \rankthree{0.852} & \rankone{0.198} & \ranktwo{25.05} & \rankthree{0.797} & \rankone{0.260} & 29.20 & \ranktwo{0.846} & \ranktwo{0.274} \\
 & SeaSplat & 25.45 & 0.794 & 0.360 & 26.44 & 0.900 & 0.322 & 20.54 & 0.808 & 0.310 & 22.58 & 0.733 & 0.400 & 23.75 & 0.809 & 0.348 \\
 & WaterSplatting & 30.86 & 0.824 & 0.273 & \rankthree{32.36} & \rankthree{0.922} & \rankthree{0.263} & 30.49 & 0.843 & 0.249 & 24.47 & 0.774 & 0.324 & \rankthree{29.55} & \rankthree{0.841} & \rankthree{0.277} \\
\cdashline{2-17}
 & \textbf{Ours} (Static) & \rankone{33.11} & \rankone{0.857} & \rankone{0.219} & \ranktwo{33.21} & \ranktwo{0.931} & \rankone{0.239} & \rankone{32.39} & \rankone{0.876} & \ranktwo{0.207} & \rankthree{25.00} & \ranktwo{0.813} & \rankthree{0.314} & \ranktwo{30.92} & \rankone{0.869} & \rankone{0.245} \\
 & \textbf{Ours} (Dynamic) & \ranktwo{32.95} & \ranktwo{0.855} & \ranktwo{0.221} & \rankone{33.48} & \rankone{0.933} & \ranktwo{0.245} & \ranktwo{32.27} & \ranktwo{0.873} & \rankthree{0.208} & \rankone{25.37} & \rankone{0.814} & \ranktwo{0.308} & \rankone{31.02} & \rankone{0.869} & \rankone{0.245} \\
\hline
\end{tabular}
}
}%
\end{table*}

\subsection{Evaluation of medium-aware SfM geometry}

\noindent \textbf{Depth and pose estimation.}
Tab.~\ref{tab:swimm3r_flsea} shows that Swimm3R achieves the best depth and pose performance across the three FLSea test sequences. Compared with the strongest baseline, MASt3R-SfM, medium-aware SfM reduces AbsRel from $0.2247$ to $0.2132$ and RMSE$_{\log}$ from $0.2006$ to $0.1915$, corresponding to relative gains of $5.1\%$ and $4.5\%$, while improving $\delta<1.25$ by $1.45$ percentage points. For pose, our model improves over MASt3R-SfM by $1.72$, $1.62$, and $2.84$ percentage points in RRA@15, RTA@15, and mAA(30), respectively, and gives larger margins over Dark3R and VGGT. On FLSea, the weaker performance of Spann3R and VGGT suggests that global feed-forward inference is less robust underwater, motivating an underwater-adapted SfM prior.

\noindent \textbf{SfM 3D map quality.}
Tab.~\ref{tab:sfm_metrics} evaluates map quality on the Barbados sequences,
where ground-truth geometry is unavailable.
COLMAP~\cite{schonberger2016structure} registers only $87.7\%$ of the frames and
yields the largest epipolar error, whereas GLUEMAP~\cite{pan2026global},
MASt3R-SfM~\cite{duisterhof2025mast3r}, and our medium-aware SfM register every
frame.
Swimm3R produces the densest map with $2.19$M points, $2.2\times$ as many as the
dense MASt3R-SfM baseline and over an order of magnitude beyond the sparse
GLUEMAP reconstruction, while simultaneously attaining the lowest
epipolar error ($0.667$) and the highest $1$-px inlier ratio ($71.5\%$).
Density and correspondence quality thus improve together rather than trading
off: against MASt3R-SfM, the only other dense baseline and hence the
like-for-like comparison, our maps are both larger and geometrically more
consistent, whereas the point counts of COLMAP and GLUEMAP mainly reflect their
sparse-by-design formulation.
Fig.~\ref{fig:point_map_comparison} shows the corresponding qualitative behavior: COLMAP and MASt3R-SfM degrade under highly turbid conditions such as the Cyan and Murky
scenes, and GLUEMAP still produces discontinuous point distributions, whereas
our medium-aware SfM yields continuous point distributions with restored point
colors.

\subsection{Evaluation of the underwater Beta-splatting map}
\noindent \textbf{Rendering performance evaluation on test frames.}
Since test frames are excluded from medium-aware SfM, we follow the test-time pose initialization and pose refinement protocol of 3R-GS~\cite{huang20253r}. We refine only each test camera pose for $300$ steps using its test image while keeping the optimized scene map fixed, and apply the same protocol to every method when evaluating PSNR, SSIM, and LPIPS.

\noindent \textbf{Photometric rendering quality.}
Tab.~\ref{tab:quantitative_comparison} and Fig.~\ref{fig:render_results} compare test-frame rendering quality across SfM initializations and rendering methods. On average, our medium-aware SfM provides a stronger restored dense point cloud initialization for downstream renderers. Averaged over the three prior rendering methods including 3R-GS, SeaSplat, and WaterSplatting, replacing GLUEMAP with our model improves PSNR from $25.30$ dB to $27.50$ dB and reduces LPIPS from $0.412$ to $0.300$. Averaged over 3R-GS, SeaSplat, and WaterSplatting, replacing MASt3R-SfM initialization with Swimm3R improves PSNR by $1.10$ dB and reduces LPIPS by
$19.7\%$.

UWBS improves rendering quality on average across all SfM initializations, with PSNR gains of $1.43$ dB, $2.77$ dB, and $1.47$ dB over the strongest non-UWBS renderer under GLUEMAP, MASt3R-SfM, and Swimm3R initialization, respectively. It also achieves the best average SSIM and LPIPS within each initialization group.
With Swimm3R initialization, the static and dynamic UWBS variants perform comparably, and the better of the two leads every scene in PSNR and SSIM.
The Caustic sequence remains challenging because rapidly varying non-Lambertian illumination is difficult to explain with a static medium field, as also noted by caustic-aware methods such as RecGS~\cite{zhang2024recgs}. Qualitatively, Fig.~\ref{fig:render_results} shows that UWBS better preserves seafloor structure, texture, and medium boundaries under challenging underwater conditions.

\begin{figure*}[t]
\centering
\setlength{\tabcolsep}{1.5pt}
\resizebox{0.9\textwidth}{!}{%
\begin{tabular}{@{}M{0.24\textwidth} M{0.24\textwidth} M{0.24\textwidth} M{0.24\textwidth}@{}}
{\small 3R-GS} & {\small SeaSplat} & {\small WaterSplatting} & {\small \textbf{UWBS}} \\
\includegraphics[width=0.24\textwidth]{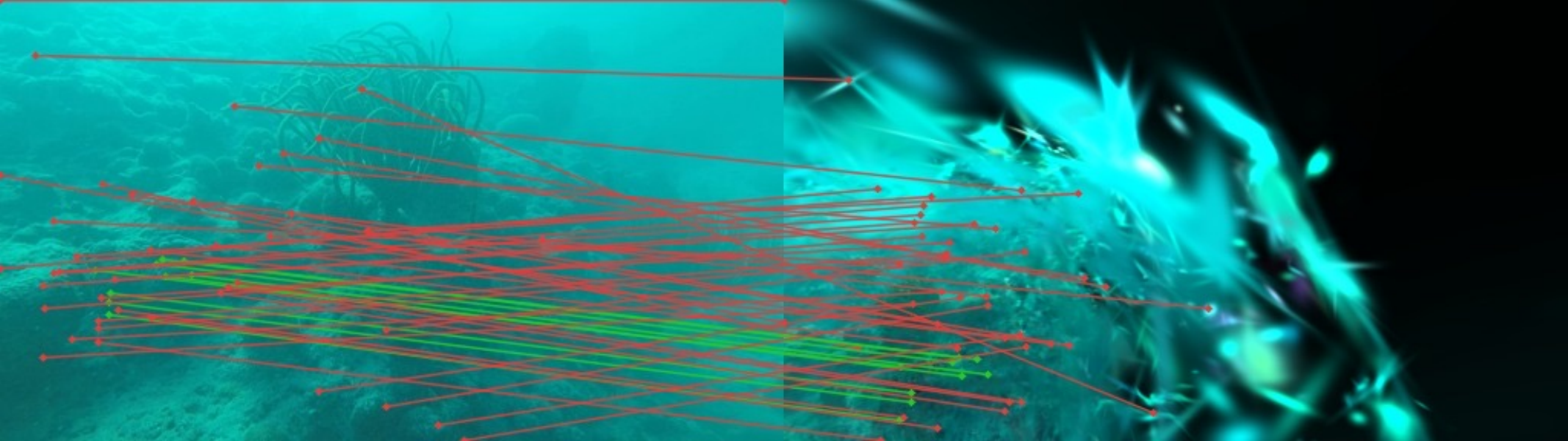}
 & \includegraphics[width=0.24\textwidth]{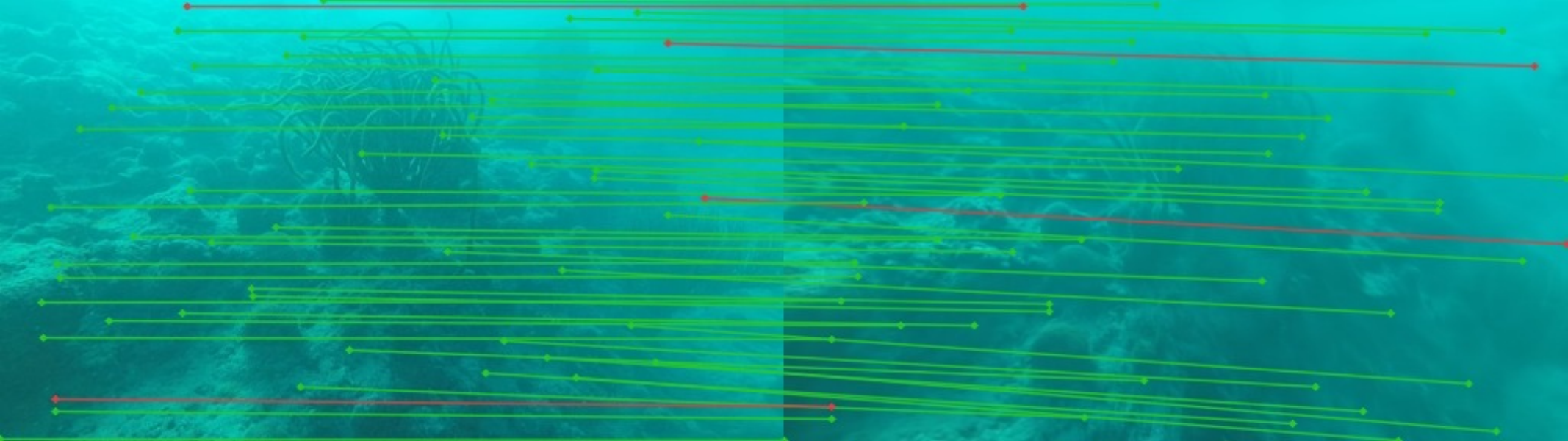}
 & \includegraphics[width=0.24\textwidth]{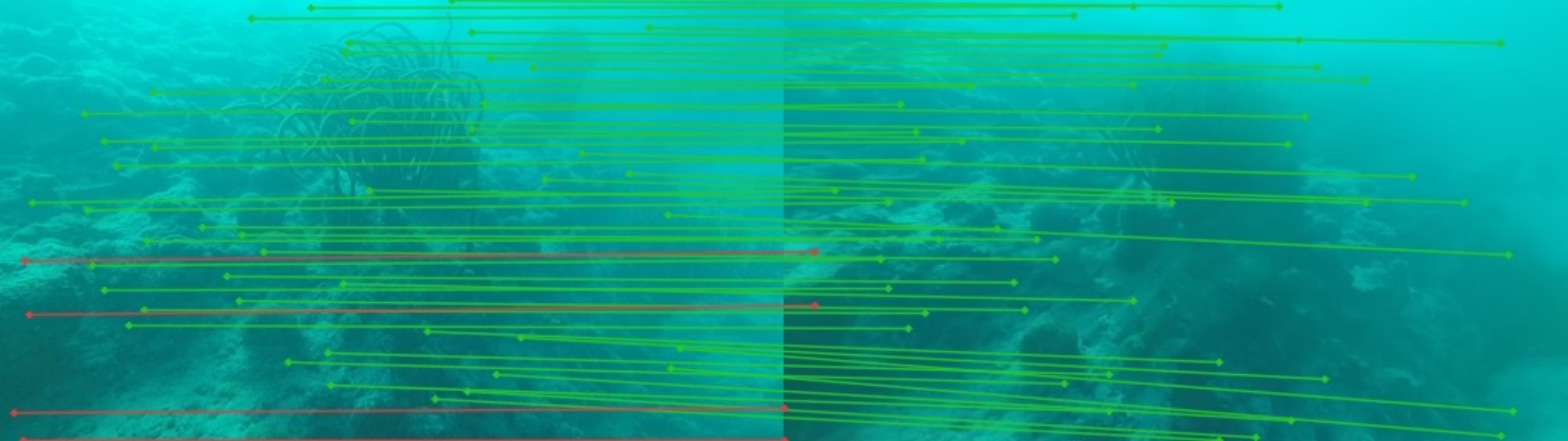}
 & \includegraphics[width=0.24\textwidth]{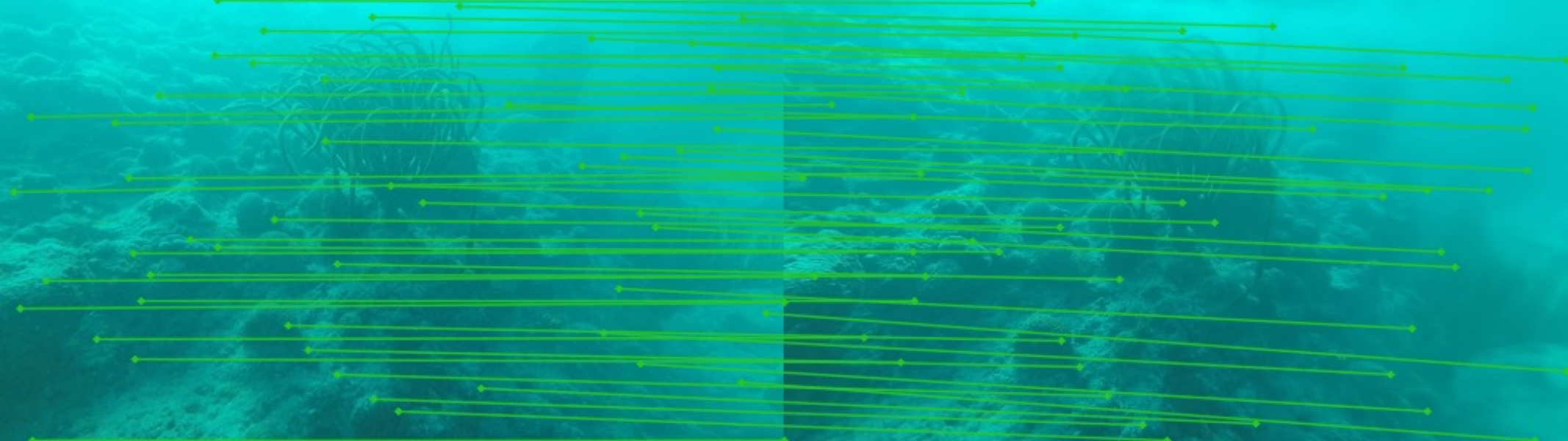} \\
\includegraphics[width=0.24\textwidth]{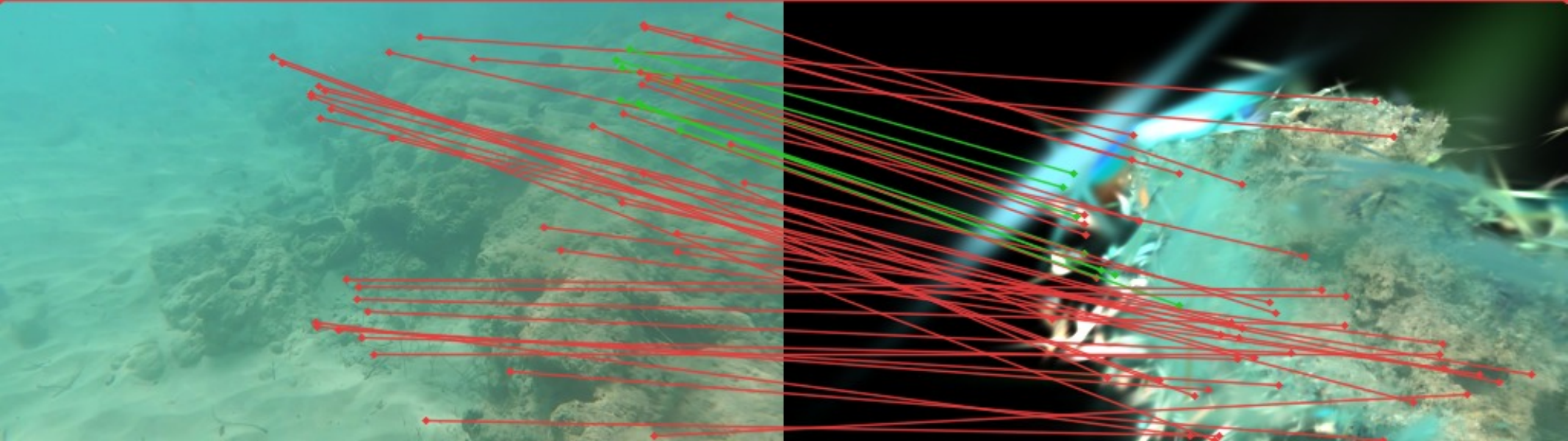}
 & \includegraphics[width=0.24\textwidth]{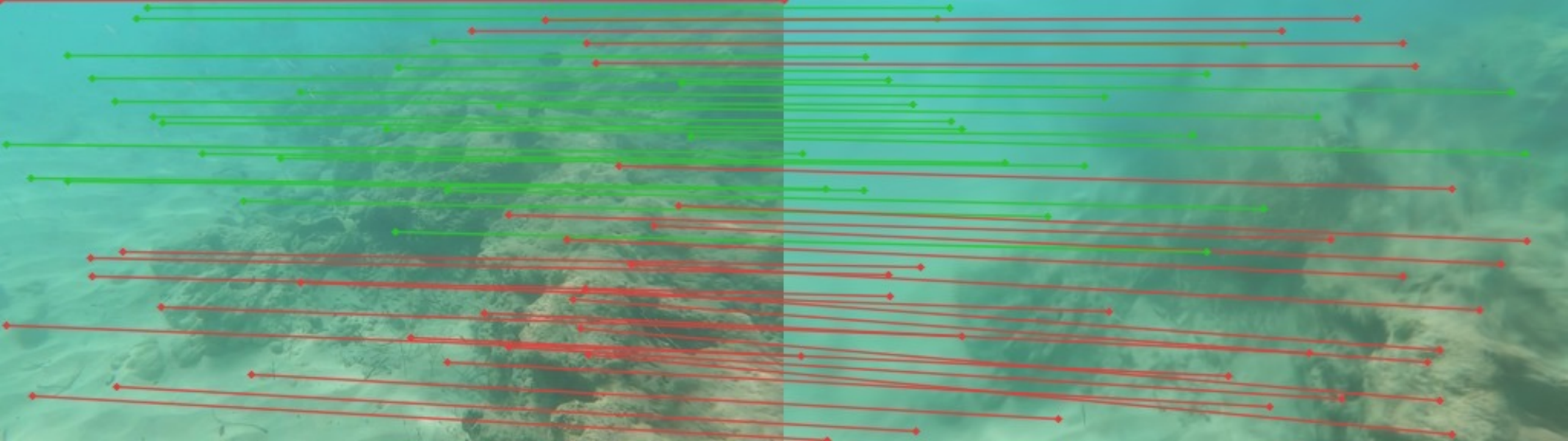}
 & \includegraphics[width=0.24\textwidth]{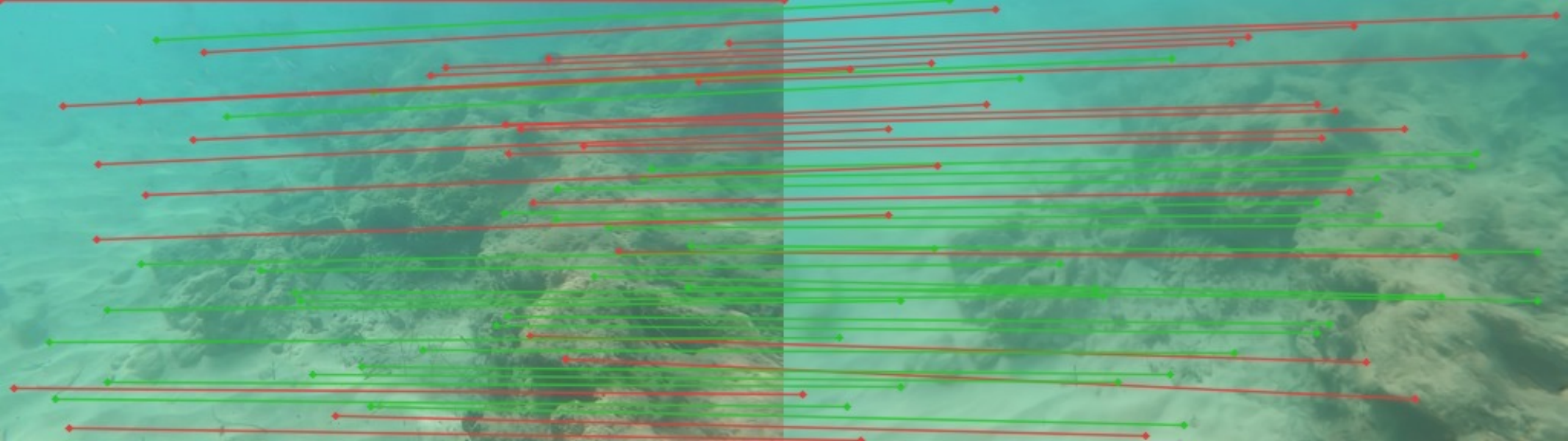}
 & \includegraphics[width=0.24\textwidth]{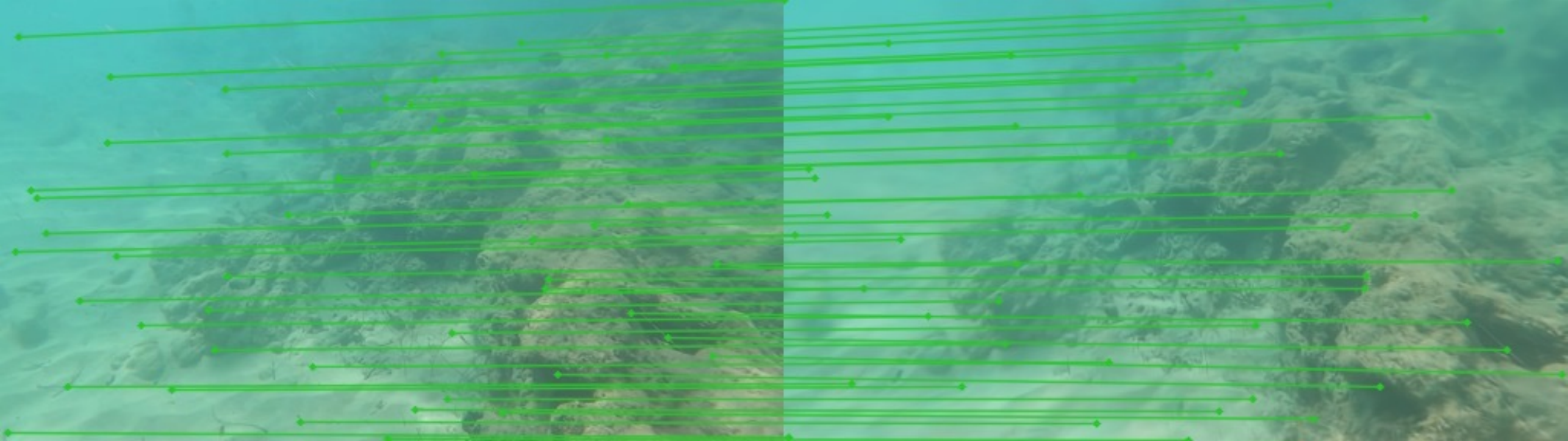} \\
\includegraphics[width=0.24\textwidth]{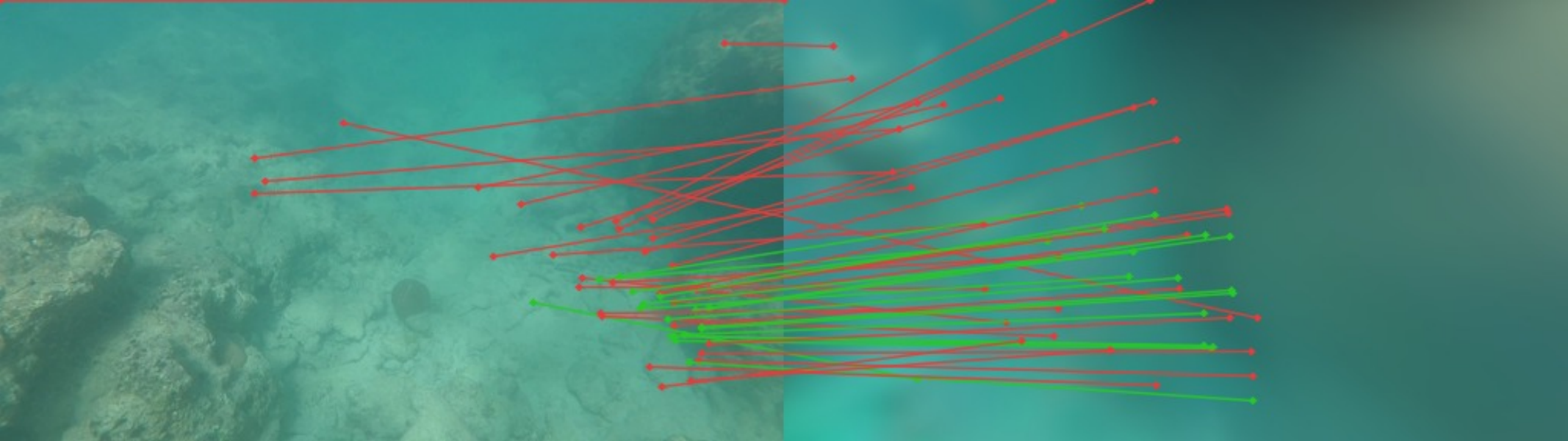}
 & \includegraphics[width=0.24\textwidth]{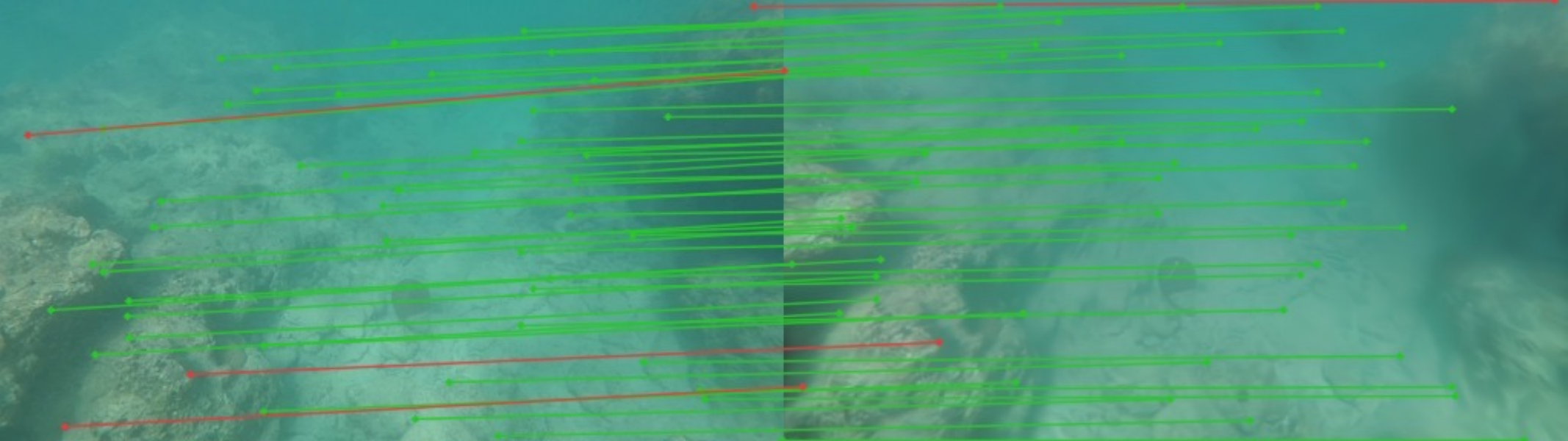}
 & \includegraphics[width=0.24\textwidth]{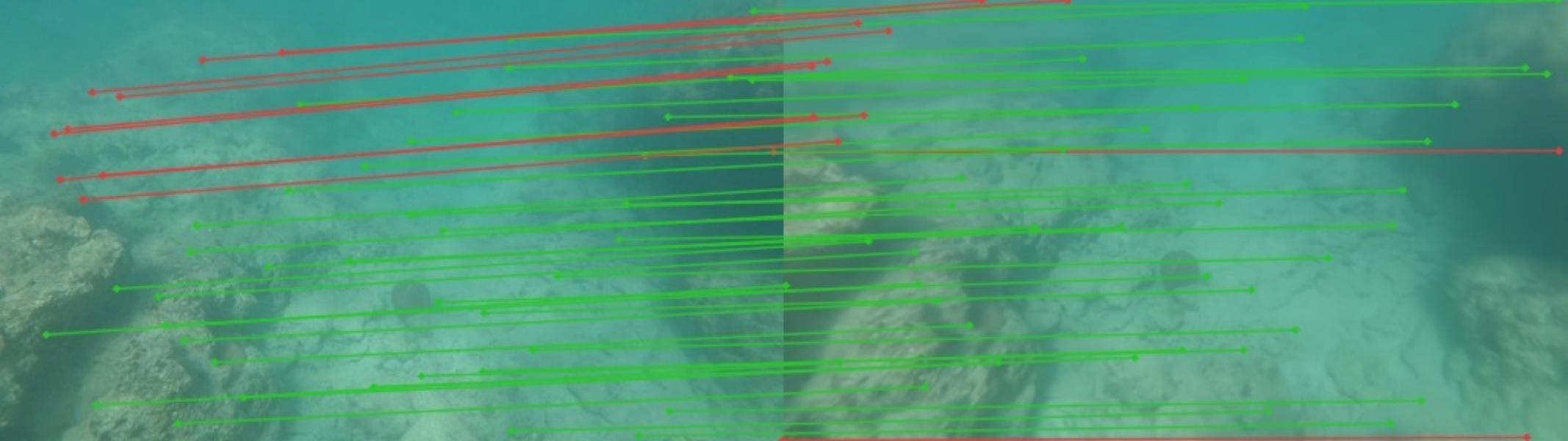}
 & \includegraphics[width=0.24\textwidth]{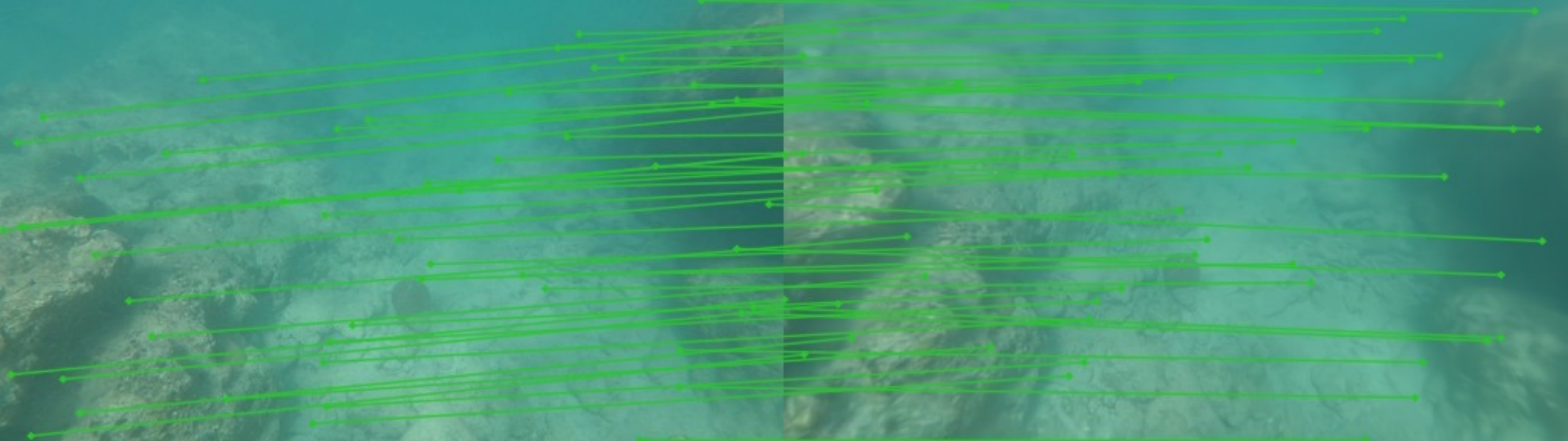} \\
\end{tabular}%
}
\caption{Qualitative query-to-render feature matching for visual localization. UWBS produces more coherent matches across underwater appearance changes, showing that its rendered maps provide more reliable 2D correspondences and render-depth for downstream PnP estimation.}
\label{fig:matching}
\end{figure*}

\noindent \textbf{Underwater visual localization.}
Tab.~\ref{tab:visloc_base} evaluates downstream map-based localization on the Barbados test frames using Swimm3R-initialized maps with different rendering methods. UWBS achieves the best pose-estimation performance with competitive runtime. Compared with WaterSplatting, UWBS improves the $5$-px inlier ratio by $9.4$ percentage points, reduces reprojection error by $0.43$, and improves RRA@15 and RTA@15 by $2.0$ and $2.4$ percentage points, respectively.

Although UWBS has the lowest rendering speed among the underwater-aware renderers, it still maintains real-time rendering at $192$ FPS, making rendering a minor cost in the overall localization pipeline. Compared with WaterSplatting, UWBS increases end-to-end localization throughput from $0.83$ FPS to $0.97$ FPS despite its lower rendering FPS. These gains are consistent with more reliable query-to-render matching and rendered depth for PnP, rather than increased memory usage. As shown in Fig.~\ref{fig:matching}, UWBS yields more coherent query-to-render matches, whereas 3R-GS suffers from increased appearance mismatch due to the lack of underwater medium modeling.

\begin{table}[t]
\centering
\caption{Underwater visual localization performance of different rendering methods using maps initialized from Swimm3R on the Barbados dataset.}
\label{tab:visloc_base}
\setlength{\tabcolsep}{3pt}
\resizebox{\linewidth}{!}{%
\begin{tabular}{l|cccc @{\hskip 1.2em} ccc}Method & Inlier@5px\,(\%)\,$\uparrow$ & Reproj\,$\downarrow$ & RRA@15\,$\uparrow$ & RTA@15\,$\uparrow$ & Ren.\ FPS\,$\uparrow$ & Loc.\ FPS\,$\uparrow$ & VRAM (GB)\,$\downarrow$ \\
\hline
\textbf{UWBS (Ours)}
& \rankone{93.9} & \rankone{1.23} & \rankone{85.2} & \rankone{96.3} & 192 & \rankone{0.97} & \rankthree{3.55} \\
WaterSplatting
& \ranktwo{84.5} & \ranktwo{1.66} & \ranktwo{83.2} & \ranktwo{93.9} & \rankone{461} & \rankthree{0.83} & \ranktwo{3.47} \\
SeaSplat
& \rankthree{83.9} & \ranktwo{1.66} & \ranktwo{83.2} & \rankthree{93.2} & \ranktwo{445} & \ranktwo{0.96} & \rankone{3.43} \\
3R-GS
& 24.2 & \rankthree{2.67} & \rankthree{21.5} & 5.8 & \rankthree{289} & 0.79 & 3.63 \\
\hline
\end{tabular}
}%
\end{table}

\begin{table}[t]
\centering
\caption{Effectiveness of the proposed SfM components on the FLSea dataset. \textit{Distillation-only} disables all physics
losses and parametric water augmentation.}
\label{tab:swimm3r_flsea_ablation}
\resizebox{\linewidth}{!}{%
{\Large
\begin{tabular}{l|ccc|ccc}
 & \multicolumn{3}{c|}{\textit{Depth}}
 & \multicolumn{3}{c}{\textit{Pose}} \\
Configuration
& AbsRel$\downarrow$ & $\delta\!<\!1.25\uparrow$ & RMSE$_{\log}\downarrow$
& RRA@15$\uparrow$ & RTA@15$\uparrow$ & mAA(30)$\uparrow$ \\
\hline
\textbf{Swimm3R (full)}
& \rankone{0.2132} & \rankone{95.11} & \rankone{0.1915}
& \ranktwo{79.19} & \ranktwo{88.43} & \rankone{69.67} \\
\quad w/o $\mathcal{L}_{\mathrm{for}}$
& \ranktwo{0.2135} & \ranktwo{94.96} & \rankone{0.1915}
& 78.93 & 86.95 & \rankthree{69.07} \\
\quad w/o $\mathcal{L}_{\mathrm{inv}}$
& 0.2179 & 94.44 & \ranktwo{0.1936}
& 77.73 & 85.38 & 66.84 \\
\quad w/o $\mathcal{L}_{\mathrm{sat}}$
& 0.2188 & 94.49 & 0.1957
& 78.03 & \rankthree{88.23} & \ranktwo{69.34} \\
\quad w/o water aug.
& \rankthree{0.2175} & \rankthree{94.49} & \rankthree{0.1945}
& \rankone{79.59} & \rankone{89.16} & 67.32 \\
\hline
\textit{Distillation-only}
& 0.2209 & 94.31 & 0.1977
& \rankthree{79.19} & 86.76 & 64.11 \\
\hline
\end{tabular}
}
}%
\end{table}

\subsection{Ablation Studies}
Tab.~\ref{tab:swimm3r_flsea_ablation} evaluates the contribution of each medium-aware SfM component. The full model achieves the best depth accuracy and the highest mAA at $30^\circ$. Removing the full physics stack causes the largest overall degradation, reducing mAA from $69.67$ to $64.11$ and increasing AbsRel from $0.2132$ to $0.2209$. This shows that the proposed physics-aware objective is important for learning underwater geometry. Removing $\mathcal{L}_{\mathrm{sat}}$ further increases AbsRel from $0.2132$ to $0.2188$ and RMSE$_{\log}$ from $0.1915$ to $0.1957$, showing that the saturation constraint also stabilizes the physics-aware training objective.

Tab.~\ref{tab:barbados_color_noref} further evaluates restored radiance quality on the Barbados using restored image quality metrics. The full model improves all three reference-free restoration metrics, UIQM, UCIQE, and URank~\cite{guo2023uranker}, over the evaluated ablations. Although removing water augmentation slightly improves FLSea localization, it substantially degrades Barbados restoration, underscoring its role in robust underwater appearance modeling.

Tab.~\ref{tab:ablation_compact} ablates the UWBS module. SGG reduces LPIPS from $0.260$ to $0.245$, indicating sharper perceptual reconstruction under medium effects. Adding depth regularization further improves PSNR and SSIM, while the best LPIPS is retained by SGG and the full model. The full model achieves the best PSNR and SSIM while matching the best LPIPS.

\begin{table}[t]
\centering
\caption{SfM restoration quality on the Barbados dataset.}
\label{tab:barbados_color_noref}
\setlength{\tabcolsep}{5pt}
\renewcommand{\arraystretch}{1.15}
{\scriptsize
\begin{tabular}{l|ccc}
Configuration & UIQM$\uparrow$ & UCIQE$\uparrow$ & URank$\uparrow$ \\
\hline
\textbf{Swimm3R (Full)}                 & \rankone{2.675} & \rankone{26.10} & \rankone{0.619} \\
w/o $\mathcal{L}_{\mathrm{for}}$        & 2.341 & 18.75 & $-0.043$ \\
w/o $\mathcal{L}_{\mathrm{inv}}$        & \ranktwo{2.569} & \rankthree{25.06} & \rankthree{0.421} \\
w/o $\mathcal{L}_{\mathrm{sat}}$        & \rankthree{2.492} & \ranktwo{25.19} & \ranktwo{0.600} \\
w/o water aug.                          & 2.264 & 21.39 & $-0.233$ \\
\hline
\textit{Distillation-only}              & 1.812 & 13.62 & $-0.595$ \\
\hline
\end{tabular}
}
\end{table}

\begin{table}[t]
\centering
\caption{Effectiveness of the proposed UWBS method on the Barbados dataset.}
\label{tab:ablation_compact}
\setlength{\tabcolsep}{5pt}
\renewcommand{\arraystretch}{1.15}
{\scriptsize
\begin{tabular}{l|ccc}
Method & PSNR$\uparrow$ & SSIM$\uparrow$ & LPIPS$\downarrow$ \\
\hline
Baseline & 30.76 & 0.864 & \rankthree{0.260} \\
+ SGG & \rankthree{30.86} & \rankthree{0.866} & \rankone{0.245} \\
+ Depth reg. & \ranktwo{30.95} & \ranktwo{0.868} & \ranktwo{0.256} \\
\textbf{UWBS (full)} & \rankone{31.02} & \rankone{0.869} & \rankone{0.245} \\
\hline
\end{tabular}
}
\end{table}

\section{Conclusion}
We presented Swimm3R, which adapts MASt3R-SfM to underwater imagery and couples it with UWBS. From scattering-degraded images it predicts restored dense point clouds, camera poses, and optical parameters that guide the joint optimization of Beta primitives and an explicit medium field, aided by Scattering-aware Geometric Gradients and edge-aware inverse-depth regularization against geometry--medium entanglement. On FLSea and Barbados, Swimm3R improves geometry estimation, test-view rendering, and downstream localization under turbid conditions.

\noindent \textbf{Limitations and Future Work.}
Since metric ground-truth poses are unavailable for these sequences, we evaluate against COLMAP and GLUEMAP references following common practice, so our pose numbers are comparative rather than exact. Our localization study likewise varies only the renderer, with the SfM side evaluated separately in Tab.~\ref{tab:sfm_metrics}. Beta primitives also render slower than 3D Gaussian primitives, a limitation shared with UBS. Future work will improve rendering efficiency, tabulate SfM-varied localization, and extend the framework toward underwater SLAM.

\bibliographystyle{IEEEtran}
\bibliography{ref}
\end{document}